\documentclass[10pt,twocolumn,letterpaper]{article}

\usepackage[pagenumbers]{cvpr} 

\usepackage{indentfirst}
\usepackage{amsmath}
\usepackage{mathtools}
\usepackage{tabularx}
\usepackage{booktabs}
\usepackage{graphicx}
\usepackage{array}
\usepackage{xcolor}
\usepackage{float}
\usepackage{colortbl}
\usepackage{multirow}
\usepackage[accsupp]{axessibility}  

\definecolor{cvprblue}{rgb}{0.21,0.49,0.74}
\usepackage[pagebackref,breaklinks,colorlinks,allcolors=cvprblue]{hyperref}

\definecolor{cosmiclatte}{rgb}{1.0, 0.97, 0.91}

\newcommand{\template}{\mathcal{I}_{T}}
\newcommand{\query}{\mathcal{I}_{Q}}
\newcommand{\rendered}{\mathcal{I}_{R}}
\newcommand{\rendereddepth}{\mathcal{D}_{R}}

\newcommand{\feattemp}{\mathcal{F}_{T}}
\newcommand{\featquery}{\mathcal{F}_{Q}}

\newcommand{\clstensor}{\mathcal{C}}
\newcommand{\offsettensor}{\mathcal{U}}
\newcommand{\conftensor}{\mathcal{W}}

\DeclarePairedDelimiter\floor{\lfloor}{\rfloor}

\newcommand{\coarsematch}{\mathcal{M}}

\newcolumntype{Y}{>{\centering\arraybackslash}X}
\newcolumntype{R}{>{\raggedleft\arraybackslash}X}
\newcolumntype{L}{>{\raggedright\arraybackslash}X}

\makeatletter
\newcommand\newtag[2]{#1\def\@currentlabel{#1}\label{#2}}
\makeatother

\newcommand{\myvspace}{2pt}

\definecolor{lightergray}{rgb}{0.935, 0.935, 0.935}
\newcommand*{\Rowcolor}[2][\tabcolsep]{%
    \ifx\relax#1\relax\else
        \kern-\the\dimexpr#1\relax
    \fi
    \makebox[0pt][l]{%
        \fboxsep=0pt
        \colorbox{#2}{%
            \strut\kern\qrr@dimen@
        }%
    }%
    \ifx\relax#1\relax\else
        \kern\the\dimexpr#1\relax
    \fi
    \ignorespaces
}

\def\paperID{3473} 
\def\confName{CVPR}
\def\confYear{2025}

\title{Co-op: Correspondence-based Novel Object Pose Estimation}

\author{Sungphill Moon \qquad Hyeontae Son \qquad Dongcheol Hur \qquad Sangwook Kim\\
NAVER LABS\\
{\tt\small \{sungphill.moon, son.ht, dongcheol.hur, o.s.w.a.l.k\}@naverlabs.com}
}

\begin{document}

\maketitle

\begin{abstract}

We propose Co-op, a novel method for accurately and robustly estimating the 6DoF pose of objects unseen during training from a single RGB image.
Our method requires only the CAD model of the target object and can precisely estimate its pose without any additional fine-tuning.
While existing model-based methods suffer from inefficiency due to using a large number of templates, our method enables fast and accurate estimation with a small number of templates.
This improvement is achieved by finding semi-dense correspondences between the input image and the pre-rendered templates.
Our method achieves strong generalization performance by leveraging a hybrid representation that combines patch-level classification and offset regression.
Additionally, our pose refinement model estimates probabilistic flow between the input image and the rendered image, refining the initial estimate to an accurate pose using a differentiable PnP layer.
We demonstrate that our method not only estimates object poses rapidly but also outperforms existing methods by a large margin on the seven core datasets of the BOP Challenge, achieving state-of-the-art accuracy.

\end{abstract}    
\section{Introduction}
\label{sec:intro}

\begin{figure}[t]
	\centering
	\includegraphics[width=\linewidth]{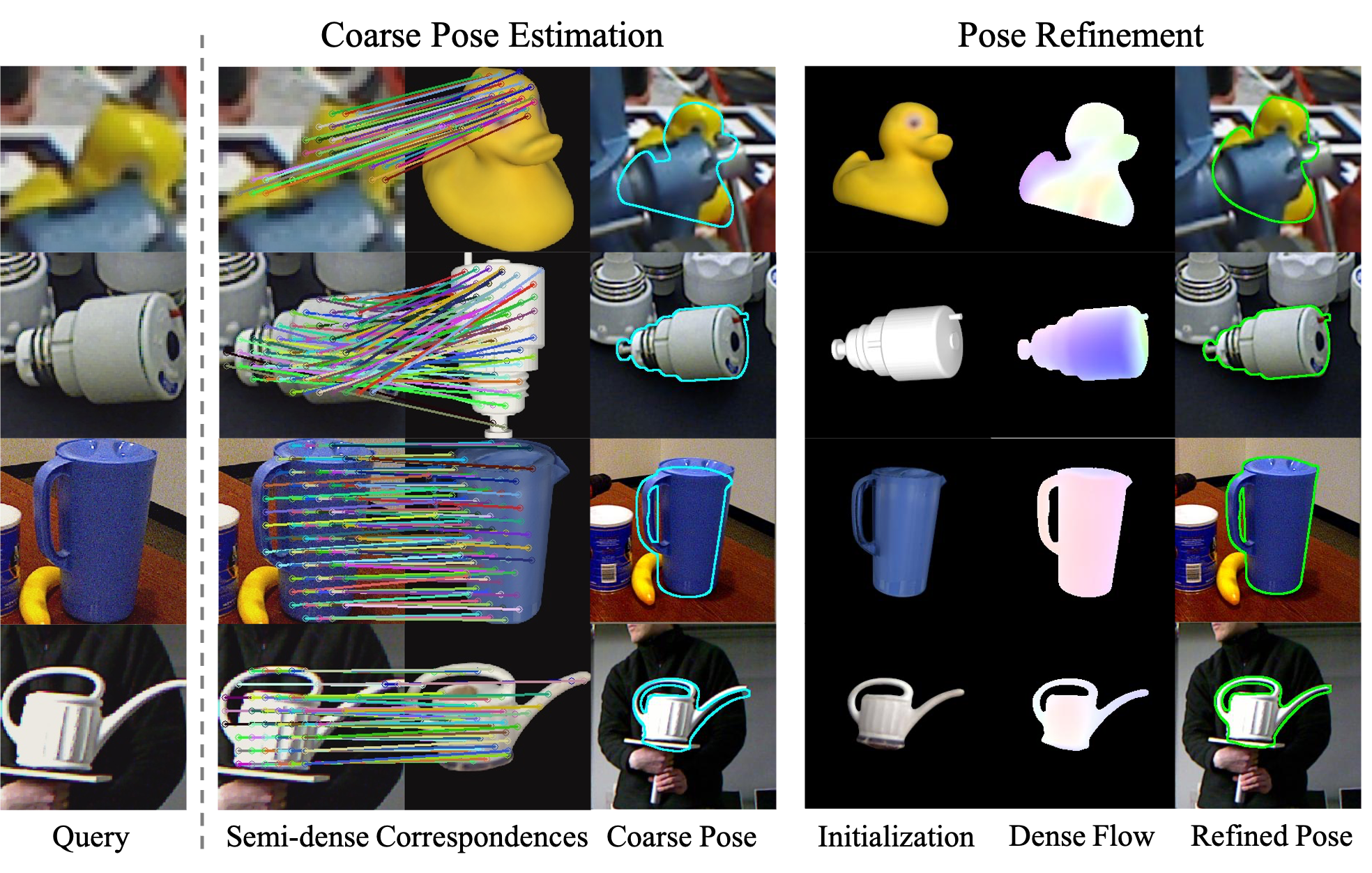}

	\caption{{\bf Examples of 6D pose estimation of novel objects.} Our method estimates semi-dense or dense correspondences between the input image and rendered images and uses them to estimate the pose.}
	\label{fig:thumbnail}
\end{figure}

Over the past decade, significant improvements in the accuracy of 6D object pose estimation have been achieved through advancements in deep learning methods \cite{xiang2018posecnn,peng2019pvnet,liu2022gdrnpp_bop,Wang_2021_GDRN,tekin18,rad2017bb8,lin2022icra:centerpose}.
However, despite their remarkable accuracy, these methods are challenging to apply in diverse, real-world environments because handling new objects requires acquiring large amounts of data and retraining the models for several hours or even days.

To address these challenges, recent studies have explored various methods.
One approach is the category-level methods \cite{lin2022icra:centerpose, Wang_2019_CVPR}, which estimate the pose of unseen objects within known categories but struggle with entirely new ones. 
Another approach aims for generalization by training on large-scale pose estimation datasets and is divided into two types: model-free and model-based methods.
However, model-free methods \cite{sun2022onepose, he2022oneposeplusplus, liu2022gen6d, li2023nerf} often face challenges with low-texture object and report results only on limited datasets or those without occlusions.

Given these limitations, model-based unseen object pose estimation methods \cite{labbe2022megapose, moon2024genflow, nguyen2024gigapose, ornek2024foundpose, wen2024foundationpose} that utilize 3D CAD models have been gaining attention.
These model-based methods typically consist of three processes—object detection, coarse pose estimation, and pose refinement—or sometimes include a fourth process, pose selection.
Since previous works like CNOS \cite{nguyen2023cnos} and SAM-6D \cite{Lin2024sam6d} have already addressed object detection with excellent generalization performance, we concentrate our research on the subsequent pose estimation process.

In this paper, we propose Co-op (\textbf{Co}rrespondence-based Novel \textbf{O}bject \textbf{P}ose Estimation), a novel method for accurately and robustly estimating the 6DoF pose of objects unseen during training from a single RGB image.
Our research begins with the consideration of how to effectively define and find correspondences between two images to achieve object pose estimation with excellent generalization performance.

By setting object pose estimation as finding correspondences between two images, our model is trained at the patch or pixel level rather than the image level.
This enables the model to learn low-level information, making it robust to domain shifts.
Furthermore, by directly leveraging the geometric and structural information of the 3D model, it enables more accurate pose estimation.
In our research, to achieve superior generalization performance, we estimate poses based on correspondences in both stages of pose estimation: coarse estimation and pose refinement.

In the first stage, coarse estimation, we estimate the initial pose of the object by estimating semi-dense correspondences between multiple templates and the input image.
In this stage, we use a hybrid representation that combines patch-level classification and offset regression to estimate the semi-dense correspondences.
Then, we estimate the pose using the Perspective-n-Point (PnP) algorithm \cite{Lepetit2009ep} based on these correspondences.
By formulating correspondence estimation not as a direct regression problem but by combining it with classification, we achieve outstanding generalization performance.
Thanks to this representation, we achieve high accuracy while using a much smaller number of templates compared to other existing coarse estimation methods.

In the second stage, pose refinement, we refine the initial pose using a render-and-compare approach.
To achieve precise refinement, we estimate probabilistic dense correspondences and learn the confidence in pose estimation by training end-to-end through a differentiable PnP layer.
This allows us to robustly estimate poses even in challenging situations such as occlusions and texture-less objects.

Our contributions can be summarized as follows:
\begin{itemize}[left=0em] 
\item We present Co-op, a novel framework for unseen object pose estimation in RGB-only cases. Co-op does not require additional training or fine-tuning for new objects and outperforms existing methods by a large margin on the seven core datasets of the BOP Challenge.
\item We propose a method for fast and accurate coarse pose estimation using a hybrid representation that combines patch-level classification and offset regression.
\item Additionally, we propose a precise object pose refinement method. It estimates dense correspondences defined by probabilistic flow and learns confidence end-to-end through a differentiable PnP layer.
\end{itemize}
\section{Related Work}
\label{sec:related_work}

\noindent \textbf{Correspondence estimation} Traditionally, finding correspondences between two images has been handled using detector-based methods involving three steps: keypoint detection, feature description, and feature matching.
These methods have been extensively researched, transitioning from handcrafted approaches \cite{Lowe2004DistinctiveIF, Bay2006surf, Calonder2012brief, Rublee2012orb} to deep learning-based approaches \cite{DeTone2018superpoint, r2d2, Dusmanu2019CVPR, tyszkiewicz2020disk}.
Recently, detector-free methods \cite{sun2021loftr, wang2024eloftr, chen2022aspanformer} have integrated feature detection and matching into a unified process, processing entire images without predefined keypoints and improving matching performance even in low-texture or repetitive areas.
Furthermore, dense matching methods \cite{edstedt2024roma, edstedt2023dkm, truong2023pdc, truong2020glu} find matches for every pixel, enhancing accuracy in tasks such as image-based localization, 3D reconstruction, and pose estimation.
Our method follows the detector-free approach in coarse estimation and pose refinement.
In the coarse estimation stage, we estimate semi-dense correspondences to achieve textureless-resilient and accurate matching, followed by the pose refinement stage, where the model estimates dense correspondences and their confidence to obtain an exact 6D pose.

\vspace*{\myvspace}
\noindent \textbf{Seen Object Pose Estimation} Deep learning-based methods have shown remarkable performance improvements with the growth of high-quality datasets and advances in GPU computing power.
Some of these methods regress sparse \cite{peng2019pvnet,tekin18,rad2017bb8, kehl2017ssd} or dense \cite{park2019pix2pose, haugaard2022surfemb, li2019cdpn} features of the object, such as keypoints or correspondences and separately recover the object’s pose by applying RANSAC \cite{FischlerB81RANSAC} and PnP \cite{Lepetit2009ep} on the regressed features.
To achieve higher accuracy, other methods use render-and-compare based refinement models \cite{li2017deepim, labbe2020,lipson2022coupled, hu2022perspective}.
Because these methods are designed to encode specific object’s features into the network weights, they tend to perform poorly when encountering unseen objects.

\begin{figure*}[t]
	\centering
	\includegraphics[width=\linewidth]{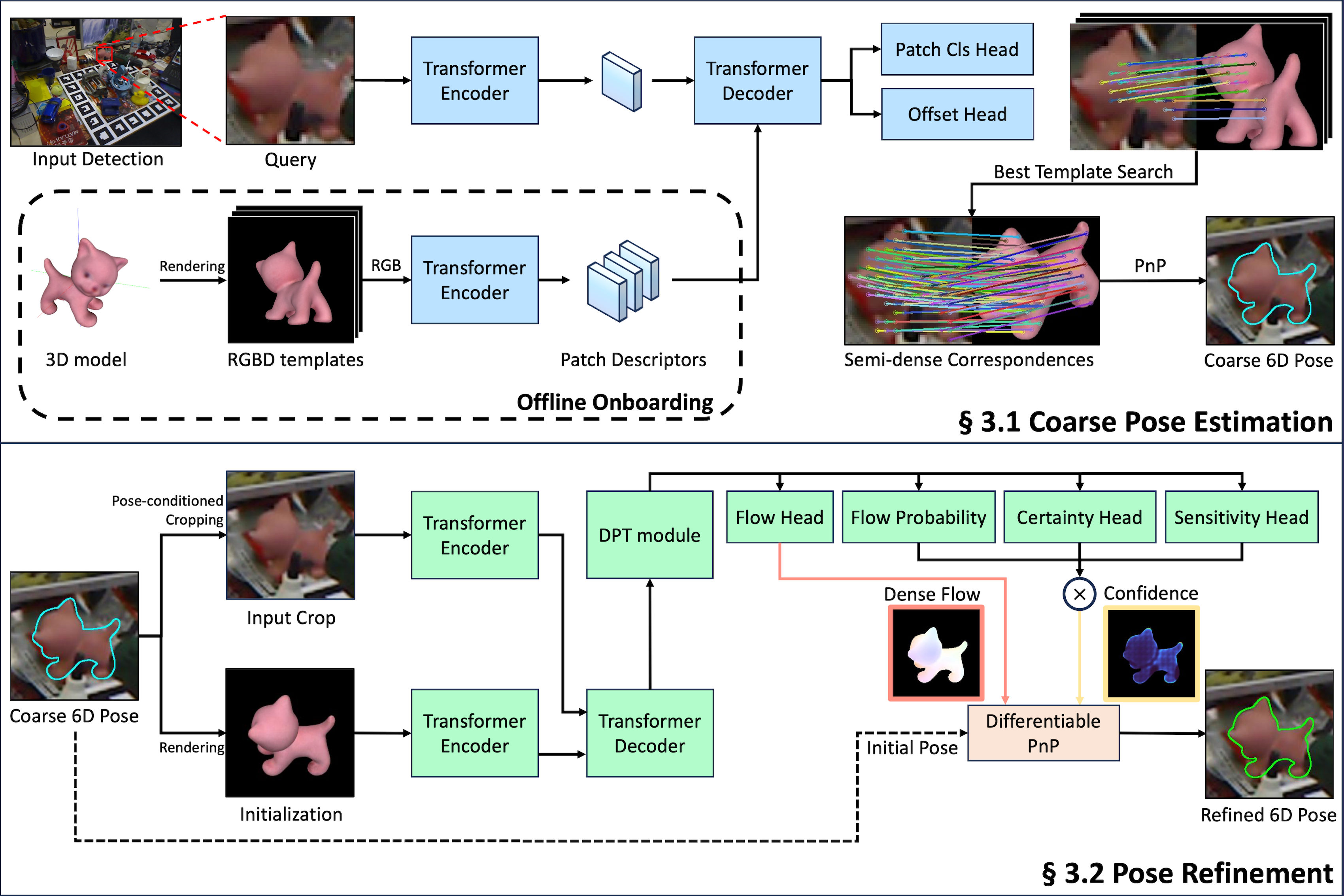}

	\caption{{\bf Overview.}
    We estimate object pose through two main stages. In the Coarse Pose Estimation stage (Sec \ref{sec:coarseestimation}), we estimate semi-dense correspondences between the query image and templates and compute the initial pose using PnP. In the Pose Refinement stage (Sec \ref{sec:poserefinement}), we refine the initial pose by estimating dense flow between the query and rendered images. Both stages utilize transformer encoders and decoders with identical structures, with the Pose Refinement stage additionally incorporating a DPT module after the decoder for dense prediction.
    }
	\label{fig:overview}
\end{figure*}

\vspace*{\myvspace}
\noindent \textbf{Unseen Object Pose Estimation} To tackle problems in more practical situations, some studies assume that objects are unknown in advance.
In particular, there has been active research on model-based unseen object pose estimation.

Some methods aim to solve the problem using template matching approaches.
These methods \cite{moon2024genflow, labbe2022megapose, okorn2021zephyr, shugurov2022osop, Nguyen2022template, wen2024foundationpose} render the 3D model to generate templates and estimate poses by selecting the most similar template to the query image.
Other methods use feature matching approaches, These methods \cite{nguyen2024gigapose, ornek2024foundpose, Nguyen2022template, Lin2024sam6d, ausserlechner2024zs6d} find 2D-3D or 3D-3D correspondences between the 3D model and the query image, recovering poses using the PnP \cite{Lepetit2009ep} algorithm.
To enhance accuracy, several methods \cite{moon2024genflow, labbe2022megapose, wen2024foundationpose} apply render-and-compare pose refinement, iteratively minimizing discrepancies between the rendered model and the query image.
Combined approaches like GenFlow \cite{moon2024genflow} and MegaPose \cite{labbe2022megapose} integrate template matching with refinement, achieving remarkable performance but incurring significant computational costs due to the need for numerous template comparisons.

To address this issue, FoundPose \cite{ornek2024foundpose} and GigaPose \cite{nguyen2024gigapose} shift from exhaustive template comparison to finding correspondences between the template and the query image, enabling faster and more robust initial pose estimation.
These methods depend on the segmentation masks from CNOS \cite{nguyen2023cnos}, to separate the object regions, and perform feature matching between the query image and the template based on DINOv2 \cite{oquab2023dinov2}.
These methods can be considered a detect-and-describe framework using the segmentation mask as a feature detector and the DINOv2 patch descriptor as a feature descriptor.

On the other hand, we propose a detector-free correspondence estimation framework for the coarse estimation stage. By not relying on segmentation, our method is robust to the noisy or partially detected regions and improves the pose accuracy.

\section{Method}
\label{sec:method}

In this section, we describe Co-op, our proposed method for unseen object pose estimation.
It consists of two main components: coarse pose estimation and pose refinement. 
Additionally, it optionally incorporates pose selection to estimate the pose more accurately. An overview of our method is presented in Fig. \ref{fig:overview}.

\subsection{Coarse Pose Estimation}
\label{sec:coarseestimation}

\noindent \textbf{Template Generation} In the coarse estimation stage, we use a set of pre-rendered templates from the 3D model.
Similar to GigaPose \cite{nguyen2024gigapose}, we minimize the number of templates required for pose estimation by generating those that include only out-of-plane rotations.

Specifically, we render the CAD model from 42 viewpoints. These viewpoints are defined by subdividing each triangle of Blender's icosphere primitive into four smaller triangles, resulting in a uniform distribution of viewpoints over the sphere, following the method used in previous works \cite{Nguyen2022template, nguyen2023cnos}.

\begin{figure}[t]
	\centering
	\includegraphics[width=\linewidth]{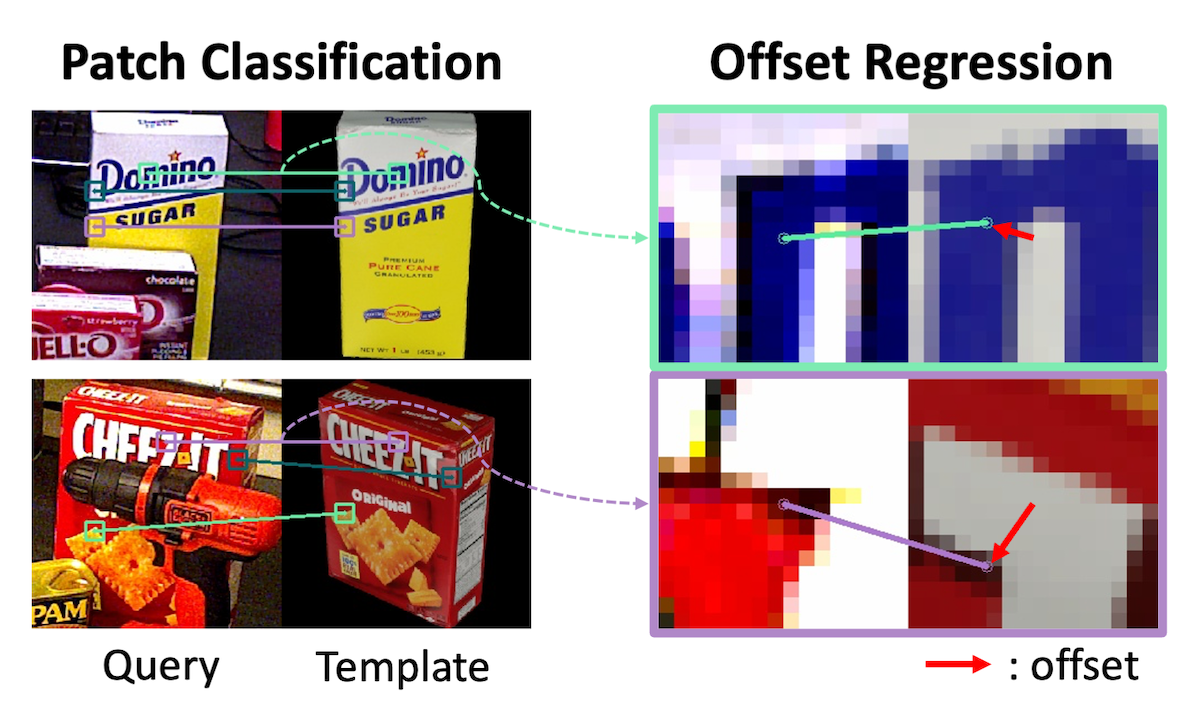}

	\caption{{\bf Visualization of Our Hybrid Representation.} Left: Patch-level classification results; matching patches are highlighted with the same color. Right: Offset regression within template patches to refine correspondences; red arrows represent the estimated offsets.}
	\label{fig:hybrid_representation}
\end{figure}

\vspace*{\myvspace}
\noindent \textbf{Hybrid Representation} The role of the coarse estimation model is to estimate semi-dense correspondences between the query image and the pre-rendered template, denoted by ${\query, \template \in \mathbb{R}^{H \times W \times 3}}$.
To enhance the generalization performance of our model, we use a hybrid representation that combines patch-level classification and offset regression.

Fig. \ref{fig:hybrid_representation} illustrates an example of our hybrid representation.
On the left, we show partial results of the patch-level classification; patches highlighted with the same color in the query image (left) and the template (right) indicate matched patches.
On the right, we show the results of offset regression within the template patches after classification, demonstrating the estimated offset from the center of the template patch that best aligns with the center of the corresponding query patch.

As shown in Fig. \ref{fig:overview}, our model consists of a Vision Transformer (ViT) encoder for extracting image features and a transformer decoder for interpreting these features.
Here, the subscript $i,j$ represents the 2D location in the feature map.
Because downsizing is performed by the encoder, each location $i,j$ in the feature map corresponds to a 16×16 patch in the input image.

The encoder takes ${\query}$ and ${\template}$ as inputs and extracts the feature maps ${\featquery, \feattemp \in \mathbb{R}^{\frac{H}{16} \times \frac{W}{16} \times 1024}}$, respectively.
The decoder and following heads then processes ${\featquery}$ and ${\feattemp}$ to compute both a tensor for patch-level classification ${\clstensor \in \mathbb{R}^{\frac{H}{16} \times \frac{W}{16} \times K}}$ and the xy-offsets ${\offsettensor \in \mathbb{R}^{\frac{H}{16} \times \frac{W}{16} \times 2}}$. 

Here, $K = \frac{H}{16} \times \frac{W}{16} + 1$ represents the number of classes for the patch-level classification. 
The first $\frac{H}{16} \times \frac{W}{16}$ classes correspond to the template patches, while the last class represents the case where there is no matching part, such as occlusion.
For each position $(i, j)$ in the feature map, $\clstensor_{i,j} \in \mathbb{R}^{K}$ contains classification scores indicating the degree of matchability between the $\frac{H}{16} \times \frac{W}{16}$ template patches and the query patch at that position.

Following this, we define the patch index as $c_{i,j} = \arg\max_{k \in {1, 2, \dots, K}} \clstensor_{i,j}^{k}$ and the offset $\offsettensor_{i,j}$ has a range of $[-0.5, 0.5]$.
At location $(i, j)$ where $c_{i,j} \neq K$, the position $M_{i,j}^{T}$ in the template corresponding to the query patch is given by:

\begin{equation}
\coarsematch_{i,j}^{T} = (
\begin{bmatrix}
		c_{i,j} \bmod 16 + 0.5 \\
		\lfloor c_{i,j} / 16 \rfloor  + 0.5
\end{bmatrix}
+ \offsettensor_{i,j} ) \times 16.
\label{eq:match}
\end{equation}

Adding $\offsettensor_{i,j}$ refines the position within the feature map grid. Multiplying by 16 maps this position back to the coordinate system of the original images $\query$ and $\template$, since each location in the feature map corresponds to a $16 \times 16$ patch in the original images.
The corresponding query image position $\coarsematch_{i,j}^Q$ is $((i + 0.5) \times 16, \, (j + 0.5) \times 16)$,
which is the center position of each patch in $\query$. The correspondence between the $\query$ and $\template$ is then defined as $\coarsematch_{i,j} = (\coarsematch_{i,j}^Q, \coarsematch_{i,j}^T)$.

\vspace*{\myvspace}
\noindent \textbf{Pose Fitting} To estimate the pose in practice, we compute the correspondences $\coarsematch$ for $\query$ and all $\template$, and select the template $k$ most similar to the query image.
The similarity score $S_{t}$ for each template $t$ is defined as follows:

\begin{equation}
S_{t} = \sum_{i,j} 
\begin{cases} 
    \max(\clstensor_{i,j}) & \text{if } c_{i,j} \neq K \\
    0 & \text{otherwise}
\end{cases}.
\end{equation}

If the predicted class $c_{i,j} = K$  (representing occlusion or unmatched patches), we exclude $\max(\clstensor_{i,j})$ from the sum.
By computing  $S_{t}$  for each template, we select the template with the highest similarity score as the best match to $\query$.
Once the most similar template is determined, we construct 2D-3D correspondences using the correspondences between the query image and the selected template along with the depth information from the template.
we estimate the initial pose by applying RANSAC \cite{FischlerB81RANSAC} with the EPnP algorithm \cite{Lepetit2009ep}.

\subsection{Pose Refinement} 
\label{sec:poserefinement}

In the refinement stage, we accurately refine the pose estimated from the coarse estimation at the pixel-level. 
To achieve this, we estimate pixel-wise dense correspondences.
As shown in Fig. \ref{fig:overview}, the structure of the refinement model is the same as the coarse estimation model with an added Dense Prediction Transformer (DPT) \cite{ranftl2021vision}.
The DPT enables pixel-wise prediction, allowing for precise pose refinement. 
Additionally, by iteratively refining the pose using a render-and-compare approach, this stage overcomes limitations that arise from using pre-rendered templates in the coarse estimation stage, such as self-occlusion.

\vspace*{\myvspace}
\noindent \textbf{Probabilistic Flow Regression} Similar to previous works \cite{moon2024genflow, hu2022perspective}, our refinement model estimates the flow between the query image $\query$ and the rendered image $\rendered$ to refine the pose.
To recover the pose accurately from the flow, it is essential to prevent inaccurate flows from significantly influencing the pose calculation.
To this end, PFA \cite{hu2022perspective} relies on RANSAC \cite{FischlerB81RANSAC} to probabilistically exclude inaccurate flows from pose estimation.
However, because RANSAC is sensitive to outlier distribution, it is insufficient when outliers are prevalent or unevenly distributed.
GenFlow \cite{moon2024genflow} focuses on reducing the influence of inaccurate flows by learning to estimate a confidence map through an end-to-end differentiable system and 6D pose loss.
Although they show accurate and robust pose estimation performance, there is still room for improvement since learning confidence does not enhance the accuracy of the flow prediction itself.

Unlike previous flow-based refinement methods \cite{hu2022perspective, moon2024genflow}, we aim to learn the conditional probability of the flow. 
In line with several correspondence estimation studies \cite{truong2023pdc, chen2022aspanformer}, we define this conditional probability as $p(Y|\query, \rendered; \theta)$, where $Y$ represents the flow between the image pair $(\query, \rendered)$, and $\theta$ denotes the network parameters.
Many methods achieve this by learning the variance of the prediction $Y$, and the predictive density is modeled using Gaussian or Laplace distributions.
We model it as a univariate Laplace distribution to simplify the problem.
Specifically, $p(Y|\query, \rendered; \theta)$ is modeled as a Laplace distribution with mean $\mu \in \mathbb{R}^{H \times W \times 2}$ and scale $b \in \mathbb{R}^{H \times W \times 1}$, both predicted by the network.
Formulating flow estimation as a probabilistic regression makes our model concentrate on highly reliable correspondences by adapting the scale parameter $b$.

\vspace*{\myvspace}
\noindent \textbf{Flow Confidence} To compute the pose using the flow estimated by the model, we need a confidence $\conftensor \in \mathbb{R}^{H \times W \times 1}$. 
This value of $\conftensor$ determines how much weight each flow error will have in calculating the pose in the differentiable PnP layer. 
$\conftensor$ is calculated from the certainty, sensitivity, and flow probability, which are learned through different loss functions.
Certainty estimates whether the flow from $\rendered$ to $\query$ is occluded or not.
Sensitivity is learned from the pose loss and highlights areas with rich textures or extremities of the object.
Similar to PDC-Net+ \cite{truong2023pdc}, we define the flow probability $P_{R}$, which represents the probability that the true flow $y$ is within a radius $R$ from the estimated mean flow $\mu$.
This can be calculated as follows:

\begin{equation}
P_{R} = P(\|y-\mu\|_1 < R) = 1 - \exp\left(-\frac{R}{b}\right).
\end{equation}

$P_R$ is an interpretable measure of reliability, representing the accuracy of the flow with a threshold $R$.
The flow confidence $\conftensor$ is calculated as the element-wise multiplication of certainty, sensitivity, and $P_R$.
This means that $\conftensor$ will have higher values when there is no occlusion (high certainty), discriminative information is available to solve the pose (high sensitivity), and the flow is accurate (high $P_R$).

\vspace*{\myvspace}
\noindent \textbf{Pose Update} To compute the refined 6D pose  $\mathbf{P}_{\text{refined}}$ using the flow $Y$ and flow confidence $\conftensor$, we employ a Levenberg-Marquardt (LM) algorithm-based PnP solver.
Given the input pose $\mathbf{P}_{\text{input}} = [\mathbf{R}_{\text{input}} \mid \mathbf{t}_{\text{input}}] $, the camera intrinsic matrix $\mathbf{K}$, and the depth map $\rendereddepth$ corresponding to $\rendered$, we calculate the 3D spatial coordinates $\mathbf{x}_{u,v}^{\text{3D}}$ corresponding to $\rendered$ as follows:

\begin{equation}
\mathbf{x}_{u,v}^{\text{3D}} = \mathbf{R}_{\text{input}}^{-1}(\mathbf{K}^{-1}\rendereddepth(u,v)\mathbf{x}_{u,v}^{\text{2D}} - \mathbf{t}_{\text{input}}),
\end{equation}

\noindent where $(u, v)$ are the pixel coordinates in $\rendered$, $\rendereddepth(u,v)$ is the depth value at each pixel, and $\mathbf{x}_{u,v}^{\text{2D}}=(u,v,1)^{T}$.
We optimize the 6D pose by minimizing the sum of squared weighted reprojection errors as follows:

\begin{align}
\begin{split}
    \operatorname*{argmin}_{\text{R}, \text{t}} \frac{1}{2} \sum_{u} \sum_{v}
    \|
    \conftensor(u,v) \times 
    (
    \pi(
    \text{R}\mathbf{x}_{u,v}^{\text{3D}}+\text{t}) \\
    -
    ((u,v)^T + Y(u,v))
    )
    \|^{2}.
\end{split}
\end{align}

Here, $\pi$ is the reprojection function that maps 3D points in camera coordinates to 2D image points using the intrinsic camera parameters $\mathbf{K}$.

Following previous work \cite{moon2024genflow, chen2022epro}, we update the pose in three iterations using the LM algorithm and further refine it to the final pose using the Gauss-Newton algorithm.

\subsection{Pose Selection}
\label{sec:poseselection}

\begin{figure}[t]
	\centering
	\includegraphics[width=\linewidth]{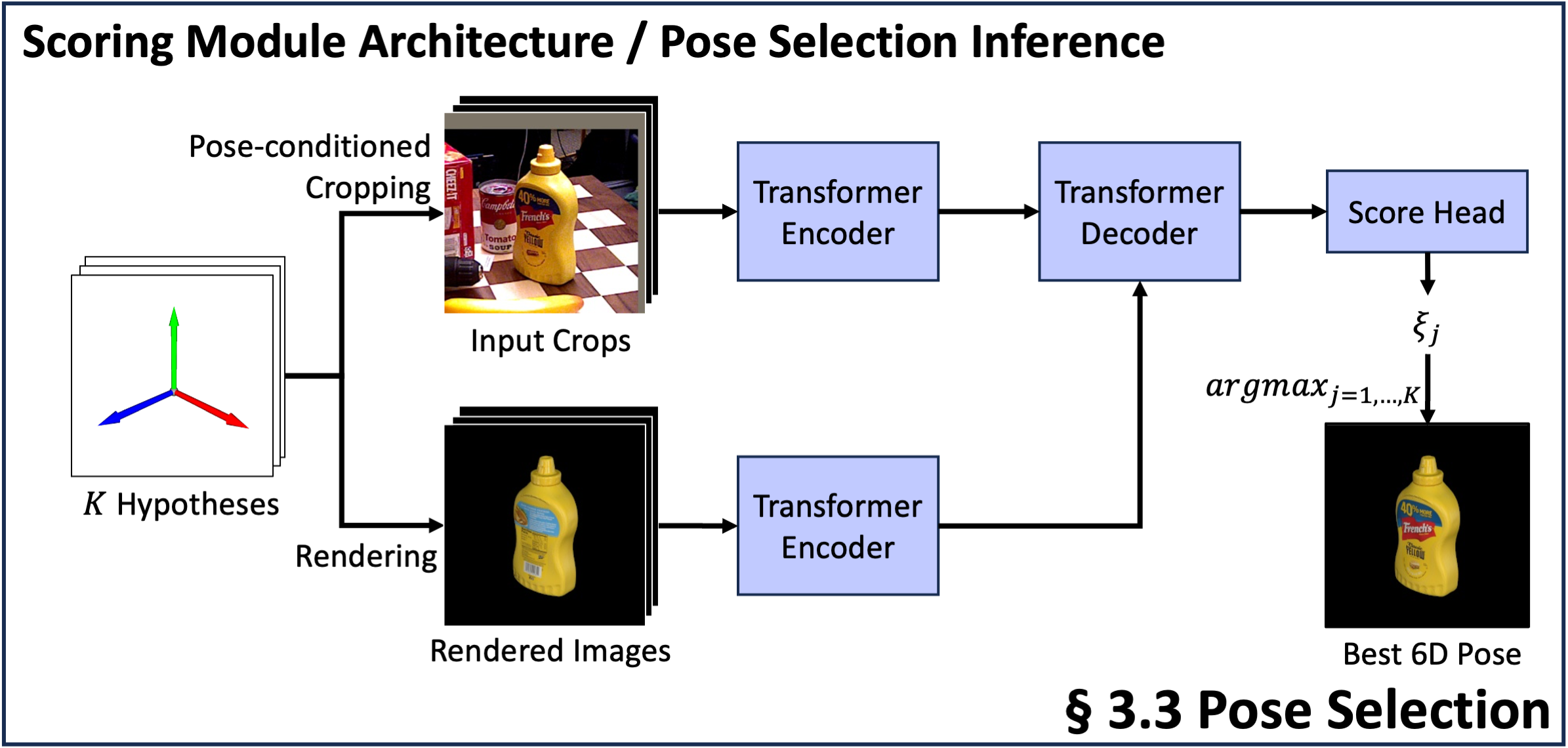}

	\caption{{\bf Pose Selection.} To achieve more precise pose estimation using a multiple hypothesis strategy, we introduce a pose selection stage (Sec \ref{sec:poseselection}).}
	\label{fig:selection}
\end{figure}

In the coarse estimation stage, the best scoring template may not provide the optimal initial pose for refinement.
For example, selecting a template rotated by 180 degrees relative to the ground truth (see Fig. \ref{fig:selection}) makes refinement challenging.
To address this issue, many methods \cite{labbe2022megapose, moon2024genflow, ornek2024foundpose, nguyen2024gigapose, wen2024foundationpose} employ a multiple hypothesis strategy: generating $N$ coarse pose estimates, refining each, and selecting the best match by comparing rendered results to the query image. 
Similar to the coarse estimation model, our pose selection model takes the query image and template as inputs but incorporates a score head to estimate a score for each pose hypothesis.
Although this increases inference time, considering multiple refined poses helps avoid cases where a difficult-to-refine template leads to a poor final pose. 

\subsection{Training}
\label{sec:training}

\noindent \textbf{Datasets} To train our three models—the coarse estimator, the pose refiner, and the pose selector—we need RGB-D images with ground-truth 6D object pose annotations.
We use the large-scale dataset provided by MegaPose \cite{labbe2022megapose}.
This dataset consists of synthetic data generated using BlenderProc \cite{Denninger2023}, featuring a diverse set of objects from ShapeNet \cite{chang2015shapenet} and Google Scanned Objects (GSO) \cite{downs2022google}, including comprehensive ground-truth 6D pose annotations and object masks.

\begin{figure}[t]
	\centering
	\includegraphics[width=\linewidth]{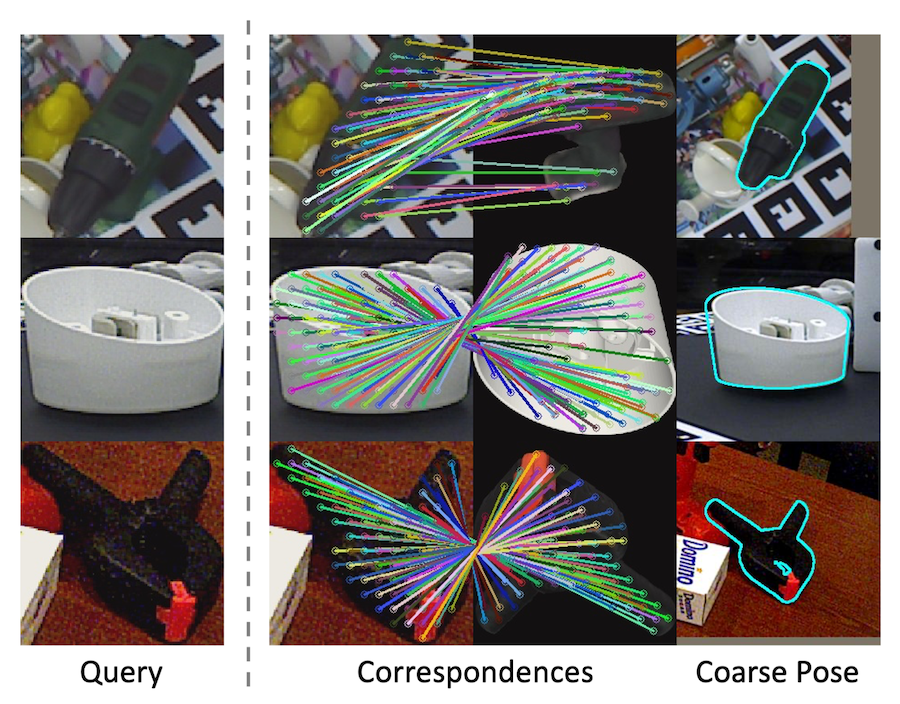}

	\caption{{
 \bf In-plane Rotation Invariant Matching Example.}
 From left to right: Query image, semi-dense correspondences between the query image and the best scoring template, and the coarse pose recovered using the PnP algorithm.}
	\label{fig:coasre_example}
\end{figure}

\vspace*{\myvspace}
\noindent \textbf{Coarse Model} Our coarse estimation model is trained to estimate the correspondence between $\query$ and $\template$. 
Similar to GigaPose \cite{nguyen2024gigapose}, we select templates that have similar out-of-plane rotations but different in-plane rotations compared to the training images cropped around the target object.
As shown in Fig. \ref{fig:coasre_example}, our model is designed to estimate correspondences between the query image and the template that are invariant to in-plane rotations.
This invariance means that the templates only need to consider out-of-plane rotations, significantly reducing the number of templates required. 
Since we have the 3D model of the target object, we can generate these correspondences by projecting the 3D model onto the image planes of both the query and template images.
Our model is trained to estimate these generated 2D-2D correspondences.

\vspace*{\myvspace}
\noindent \textbf{Refiner Model} Our refinement model is trained similarly to previous works \cite{labbe2020, labbe2022megapose, moon2024genflow} that use the render-and-compsare approach.
We generate noisy input poses $\textbf{P}_{\text{input}}$ by adding zero-mean Gaussian noise to the ground truth poses $\textbf{P}_{\text{gt}}$, with translation noise standard deviations of (0.01, 0.01, 0.05) along the x, y, and z axes, and rotation noise standard deviations of 15 degrees per axis in Euler angles.
The model is trained to predict $\textbf{P}_{\text{gt}}$ from $\textbf{P}_{\text{input}}$.

\vspace*{\myvspace}
\noindent \textbf{Selection Model} Our selection model is trained to assess the similarity between $\query$ and $\rendered$ based on $\textbf{P}_{\text{input}}$ to select the most accurate pose from multiple hypotheses.
We train the model using binary cross-entropy loss with six poses per $\textbf{P}_{\text{gt}}$: one positive, five negatives.
Positives have translation differences within (0.01, 0.01, 0.05) and rotation differences within 5 degrees of $\textbf{P}_{\text{gt}}$.
By setting small rotation thresholds for positives, we enhance the model’s ability to distinguish poses close to the ground truth, improving its discriminative capability.

\subsection{Implementation Details}
\label{sec:impledetail}

The encoder and decoder architecture of Co-op are based on CroCo \cite{weinzaepfel2023croco}, a vision foundation model trained on large-scale datasets for 3D vision tasks.
This allows us to significantly leverage the benefits of CroCo pre-training.
The coarse model processes input images of resolution $224\times224$, whereas the refinement and selection models process images of resolution $256\times256$.
For detailed information about the model configurations, learning rates, and training schedules, please refer to the supplementary material.

\vspace*{\myvspace}
\noindent \textbf{Coarse Model} In our coarse model, we define the semi-dense correspondence as a hybrid representation combining patch-level classification and offset regression.
Therefore, the coarse model is trained using two loss functions: the classification loss $\mathcal{L}_{\text{cls}}$ and the regression loss $\mathcal{L}_{\text{reg}}$.

When defining the ground-truth match from the patch center $(i, j)$ in the query image to the template as $\coarsematch_{gt}^T = (\bar{u}, \bar{v})$, the ground-truth index for $\mathcal{L}_{\text{cls}}$ is defined as $\floor*{\frac{\bar{v}}{16}}\times \left(\frac{W}{16} \right) + \floor*{\frac{\bar{u}}{16}}$ when a match exists. 
Here, $W$ is the width of the template image, and dividing by 16 accounts for the downsizing due to the patch size.
When there is no match, the ground-truth index is defined as $\left( \frac{H}{16} \times \frac{W}{16} \right) + 1$, representing an additional class for “no match.”
Therefore, $\mathcal{L}_{\text{cls}}$ is defined as the cross-entropy loss between the ground-truth index and the model’s output patch-level probability vector.

The offset ground truth $\mathcal{U}_{gt}$ for $\mathcal{L}_{\text{reg}}$ is defined as:

\begin{equation}
\mathcal{U}_{gt} = \frac{\coarsematch_{gt}^T}{16} - \floor*{\frac{\coarsematch_{gt}^T}{16}} - 0.5,
\end{equation}

\noindent where $\coarsematch_{gt}^T$ is the ground-truth match in the template.
Subtracting $\floor*{\coarsematch_{gt}^T/16}$ and 0.5 centers the offset within the patch, ranging from -0.5 to 0.5.
Here, $\offsettensor$ denotes the predicted offset output by the model.
The regression loss $\mathcal{L}_{\text{reg}}$ is then defined as the L1 loss between $\offsettensor$ and $\offsettensor_{gt}$:

\begin{equation}
\mathcal{L}_{\text{reg}} = \left\| \offsettensor - \offsettensor_{gt} \right\|_1.
\end{equation}

Therefore, the overall loss for training the coarse model, $\mathcal{L}_{\text{coarse}}$, is defined as:

\begin{equation}
\mathcal{L}_{\text{coarse}} = \mathcal{L}_{\text{cls}} + \alpha\mathcal{L}_{\text{reg}}.
\end{equation}

In the above equation, $\alpha$ is set to 2.

\vspace*{\myvspace}
\noindent \textbf{Refiner Model} Our refiner model is trained using flow loss, certainty loss, and pose loss, similar to GenFlow \cite{moon2024genflow}.
Since we parameterize the flow output of the network with a Laplace distribution, we train the flow by minimizing the negative log-likelihood. Therefore, the flow loss is defined as follows:

\begin{equation}
\mathcal{L}_{flow} = \sum_{u} \sum_{v}
\left[ 
   \frac{|\mu_{u,v} - \bar{\mu}_{u,v}|}{b_{u,v}} + 2 \log b_{u,v}
\right].
\end{equation}

In this equation, $\mu_{u,v}$ is the flow at pixel position $(u,v)$ within the rendered mask, and $\bar{\mu}_{u,v}$ is the ground-truth flow.
$b_{u,v}$  is the scale (uncertainty) at pixel position $(u,v)$.
Additionally, the certainty loss $\mathcal{L}_{cert}$ is defined as the binary cross-entropy loss that determines whether the flow from the rendered image \(\rendered\) to the query image \(\query\) falls within the ground-truth mask in \(\query\) or not.
The pose loss $\mathcal{L}_{pose}$, which quantifies the difference between the refined pose and the ground-truth pose, is defined as the distance between corresponding 3D points on the 3D model, following the approach used in previous works \cite{moon2024genflow, labbe2022megapose}.
Detailed information about $\mathcal{L}_{pose}$ is provided in the supplementary material.

The overall loss for refiner model is defined as follows:

\begin{equation}
\mathcal{L}_{refiner} = \mathcal{L}_{\text{flow}} + \beta\mathcal{L}_{\text{cert}} + \gamma\mathcal{L}_{\text{pose}}.
\end{equation}

The weights $\beta$ and $\gamma$ for each loss are set to 5 and 20, respectively.
\section{Experiments}

\setlength{\tabcolsep}{2.5pt}
\begin{table*}[t!]
    \begin{center}
    \scriptsize
    \begin{tabularx}{1.0\linewidth}{r l l |Y Y Y Y Y Y Y |Y| Y}
    \toprule
       \# \vspace{0.4ex} &
       Coarse estimation &
       Pose refinement &
       LM-O &
       T-LESS &
       TUD-L &
       IC-BIN &
       ITODD &
       HB &
       YCB-V &
       Mean &
       Time(sec)\\
    \toprule
    \multicolumn{12}{l}{\textit{Coarse estimation only:}} \\
    \midrule

    \newtag{1}{main.1} & ZS6D \cite{ausserlechner2024zs6d} & --  & 29.8 & 21.0 & -- & -- & -- & -- & {32.4} & -- & -- \\

    \newtag{2}{main.2} & MegaPose \cite{labbe2022megapose} & -- & 22.9 & 17.7 & 25.8 & 15.2 & 10.8 & 25.1 & 28.1 & 20.8 & 15.465 \\

    \newtag{3}{main.3} & GenFlow \cite{moon2024genflow} & -- & 25.0 & {21.5} & {30.0} & {16.8} & {15.4} & 28.3 & 27.7 & {23.5} & \phantom{0}3.839 \\

    \newtag{4}{main.4} & GigaPose \cite{nguyen2024gigapose} & -- & 29.9 & 27.3 & 30.2 & 23.1 & 18.8 & 34.8 & 29.0 & 27.6 & \phantom{0}\textbf{0.384} \\ 

    \newtag{5}{main.5} & FoundPose \cite{ornek2024foundpose} & -- & \underline{39.7}	& \underline{33.8}	& \underline{46.9}	& \underline{23.9}	& \underline{20.4}	& \underline{50.8}	& \underline{45.2}	& \underline{37.3}	& \phantom{0}1.690 \\

    \newtag{6}{main.6} & Co-op & -- & \textbf{59.7} & \textbf{59.2} & \textbf{64.2} & \textbf{45.8} & \textbf{39.1} & \textbf{78.1} & \textbf{62.6} & \textbf{58.4} & \phantom{0}\underline{0.979} \\
    
    \midrule
    \multicolumn{12}{l}{\textit{With pose refinement (a single hypothesis):}} \\
    \midrule

    \newtag{7}{main.7} & MegaPose \cite{labbe2022megapose} & MegaPose \cite{labbe2022megapose} & 49.9 & {47.7} & {65.3} & 36.7 & 31.5 & 65.4 & {60.1} & 50.9 & 31.724 \\

    \newtag{8}{main.8} & GenFlow \cite{moon2024genflow} & GenFlow \cite{moon2024genflow} & 54.7 & {51.4} & \underline{67.0} & 43.7 & 38.4 & 73.0 & {61.9} & 55.7 & 10.553\\

    \newtag{9}{main.9} & GigaPose \cite{nguyen2024gigapose} & MegaPose \cite{labbe2022megapose} & {55.7} & {54.1} & 58.0 & {45.0} & {37.6} & 69.3 & {63.2} & {54.7} & \phantom{0}2.301 \\

    \newtag{10}{main.10} & FoundPose \cite{ornek2024foundpose} & MegaPose \cite{labbe2022megapose} & {55.7}	& {51.0}	& {63.3}	& {43.3}	& {35.7}	& {69.7}	& \underline{66.1}	& {55.0}	& \phantom{0}6.385\\

    \newtag{11}{main.11} & GigaPose \cite{nguyen2024gigapose} & GenFlow \cite{moon2024genflow} & {59.5}	& {55.0}	& {60.7}	& {47.8}	& {41.3}	& {72.2}	& {60.8}	& {56.8}	& \phantom{0}2.213\\

    \newtag{12}{main.12} & GigaPose \cite{nguyen2024gigapose} & Co-op & \underline{60.2}	& \underline{56.0}	& {62.8}	& \underline{50.2}	& \underline{43.4}	& \underline{73.9}	& {61.9}	& \underline{58.3}	& \phantom{0}1.149\\

    \newtag{13}{main.13} & Co-op & Co-op & \textbf{64.2}	& \textbf{63.5}	& \textbf{71.7}	& \textbf{51.2}	& \textbf{47.3}	& \textbf{83.2}	& \textbf{67.0}	& \textbf{64.0}	& \phantom{0}1.852\\

    \midrule
    \multicolumn{12}{l}{\textit{With pose refinement (5 hypotheses):}} \\
    \midrule

    \newtag{14}{main.14} & MegaPose \cite{labbe2022megapose} & MegaPose \cite{labbe2022megapose} & 56.0	& 50.7	& {68.4}	& 41.4	& 33.8	& 70.4	& 62.1	& 54.7 & 47.386 \\

    \newtag{15}{main.15} & GenFlow \cite{moon2024genflow} & GenFlow \cite{moon2024genflow} & 56.3	& 52.3	& {68.4}	& 45.3	& 39.5	& \underline{73.9}	& 63.3	& 57.0 & 20.890\\

    \newtag{16}{main.16} & GigaPose \cite{nguyen2024gigapose} & MegaPose \cite{labbe2022megapose} & {59.9} & \underline{57.0} & 64.5 & {46.7} & {39.7} & 72.2 & 66.3 & {57.9} & \phantom{0}7.682\\

    \newtag{17}{main.17} &  FoundPose \cite{ornek2024foundpose} & MegaPose \cite{labbe2022megapose} & \underline{61.0}	& \underline{57.0} &	\underline{69.4}	& \underline{47.9}	& \underline{40.7}	& {72.3}	& \textbf{69.0}	& \underline{59.6}	& 20.523\\

    \newtag{18}{main.18} &  Co-op & Co-op & \textbf{65.5}	& \textbf{64.8} &	\textbf{72.9}	& \textbf{54.2}	& \textbf{49.0}	& \textbf{84.9}	& \underline{68.9}	& \textbf{65.7}	& \phantom{0}4.186\\

    \bottomrule
    \end{tabularx}
    \end{center}
    \caption{\textbf{Results on the BOP challenge datasets.} The table shows the Average Recall (AR) score for each dataset, the mean AR score across all datasets, and the time required to estimate the poses of all objects within an image. From top to bottom, the results are ordered as follows: coarse estimation only, refinement applied to a single hypothesis, and refinement results with multiple hypotheses.
    }
    \label{tab:main_results}
\end{table*}

\subsection{Experimental Setup}
\label{sec:exp_setup}

\noindent \textbf{Datasets} We evaluate our method on the seven core datasets of the BOP Challenge \cite{hodan2024bop}: LineMod Occlusion(LM-O) \cite{Brachmann2014LMO}, T-LESS \cite{hodan2017tless}, TUD-L \cite{hodan2018TUDL}, IC-BIN \cite{DoumanoglouKMK16ICBIN}, ITODD \cite{DrostUBHS17ITODD}, HomebrewedDB(HB) \cite{KaskmanZSI19HB}, and YCB-Video(YCB-V) \cite{xiang2018posecnn}.
Since our method does not require training for each dataset, we did not use any training images; only the 3D object models and test images were used.

\vspace*{\myvspace}
\noindent \textbf{Metrics} We follow the protocol of the BOP Challenge \cite{hodavn2020bop}, which consists of three pose-error functions: Visible Surface Discrepancy (VSD), Maximum Symmetry-Aware Surface Distance (MSSD), and Maximum Symmetry-Aware Projection Distance (MSPD).
The overall accuracy is measured by the Average Recall (AR), calculated as: $\text{AR} = (\text{AR}_{\text{VSD}} + \text{AR}_{\text{MSSD}} + \text{AR}_{\text{MSPD}}) \, / \, 3$.

\subsection{BOP Benchmark Results}

Table \ref{tab:main_results} shows the pose estimation results on the BOP dataset.
In our method's experiments, we use a single RTX 3090 Ti GPU.
For a fair comparison with previous works \cite{ausserlechner2024zs6d,labbe2022megapose,moon2024genflow,nguyen2024gigapose,ornek2024foundpose}, we adopt the default CNOS \cite{nguyen2023cnos} detections provided by the BOP challenge. 
We also set the number of hypotheses for the multiple hypotheses strategy to 5, consistent with the previous methods.
Under these standardized conditions, our method demonstrates improved pose estimation accuracy compared to prior approaches.
Although our method is proposed to solve the RGB-only object pose estimation problem, it is also available for RGBD input as a correspondence-based method.
The results for this can be found in the supplementary material.

\subsubsection{Coarse Estimation}

Rows \ref{main.1} to \ref{main.6} show the performance of various methods using only coarse estimation, without any refinement stage.
Our method, Co-op (row \ref{main.6}), significantly outperforms previous approaches in this setting.
Co-op achieves the highest AR score across all datasets, which is approximately 56.6\% higher than the second-best method, FoundPose \cite{ornek2024foundpose} (row \ref{main.5}).
Although GigaPose \cite{nguyen2024gigapose} is the fastest coarse estimation method with an inference time of 0.384 seconds per image (row \ref{main.4}), Co-op’s coarse estimation alone achieves a higher mean AR score than GigaPose combined with the refinement model GenFlow \cite{moon2024genflow} (row \ref{main.11}).
This demonstrates that our proposed coarse estimation method is highly effective.

\subsubsection{Pose Refinement}

Rows \ref{main.7} to \ref{main.13} present the results of combining coarse estimation and refinement using a single hypothesis.
In this setting, our method, Co-op (row \ref{main.13}), outperforms previous works \cite{moon2024genflow, labbe2022megapose}.
To focus on the performance of the refinement model independently of the coarse estimation method, we conducted experiments by combining our refinement model with GigaPose \cite{nguyen2024gigapose} (rows \ref{main.9}, \ref{main.11}, and \ref{main.12}).
This allows for a direct comparison of the refinement models using the same coarse estimation.
The results show that our refinement model consistently outperforms others, achieving higher AR scores even when paired with the same coarse estimation method.
In the multiple hypothesis experiments (rows \ref{main.14} to \ref{main.18}), our method outperforms the previous best method, FoundPose \cite{ornek2024foundpose} (row \ref{main.17}), by more than 6 points in AR score.

\def\angle{70}

\begin{table}[t!]
    \setlength{\tabcolsep}{2.5pt}
    \scriptsize %
    \begin{center}
    \begin{tabularx}{1.0\linewidth}{r l |Y Y Y Y Y Y Y |Y}
    \toprule
    \# \vspace{0.4ex} &
    Method &
    \rotatebox[origin=lB]{\angle}{LM-O} &
    \rotatebox[origin=lB]{\angle}{T-LESS} &
    \rotatebox[origin=lB]{\angle}{TUD-L} &
    \rotatebox[origin=lB]{\angle}{IC-BIN} &
    \rotatebox[origin=lB]{\angle}{ITODD} &
    \rotatebox[origin=lB]{\angle}{HB} &
    \rotatebox[origin=lB]{\angle}{YCB-V} &
    \rotatebox[origin=lB]{\angle}{Mean} \\
    \toprule
    
    & \multicolumn{9}{l}{\textit{Coarse Estimation:}} \\
    \midrule

    \newtag{1}{ab.1} & Ours (Proposed) & 59.7	& 59.2	& 64.2	& 45.8	& 39.1 & 78.1 & 62.6	& 58.4	\\
    
    \newtag{2}{ab.2} & W/o Classification & 52.6	& 56.1	& 44.1	& 40.3	& 31.0 & 71.7 & 55.8 & 50.2	\\
    
    \newtag{3}{ab.3} & W/o CroCo pretrained & 47.2	& 48.9	& 46.7	& 38.1	& 27.0 & 61.6 & 55.0	& 46.4	\\

    \midrule
    & \multicolumn{9}{l}{\textit{Pose Refinement:}} \\
    \midrule

    \newtag{4}{ab.4} & Ours (Proposed) & 64.2	& 63.5	& 71.7	& 51.2	& 47.3 & 83.2 & 67.0	& 64.0	\\
    
    \newtag{5}{ab.5} & W/o Probabilistic Flow & 63.6	& 63.2	& 71.6	& 49.5	& 46.4 & 82.8 & 65.8 & 63.3	\\

    \newtag{6}{ab.6} & W/o CroCo pretrained & 62.3	& 61.2	& 66.9	& 49.0	& 42.4 & 79.6 & 67.1	& 61.2	\\

    \midrule
    & \multicolumn{9}{l}{\textit{Pose Selection:}} \\
    \midrule

    \newtag{7}{ab.7} & Ours (Proposed) & 65.5	& 64.8	& 72.9 & 54.2	& 49.0 & 84.9 & 68.9	& 65.7	\\

    \newtag{8}{ab.8} & W/o CroCo pretrained & 65.1	& 64.4	& 72.8 & 53.1	& 47.3 & 84.2 & 68.3	& 65.0	\\
    
    \bottomrule
    \end{tabularx}
    \end{center}
    \caption{\textbf{Ablation Study.} We present the results for each stage under various settings. Rows \ref{ab.1} to \ref{ab.3} correspond to Coarse Estimation, rows \ref{ab.4} to \ref{ab.5} to Pose Refinement, and rows \ref{ab.7} and \ref{ab.8} to Pose Selection. Please refer to Section \ref{sec:ablation} for more details.
    }
    \label{tab:ablations}
\end{table}

\subsection{Ablation Study}
\label{sec:ablation}

Table \ref{tab:ablations} presents various ablation evaluations on seven core datasets, as described in Section \ref{sec:exp_setup}. 
The results for each of our stages are shown in rows \ref{ab.1}, \ref{ab.4}, \ref{ab.7}, respectively. 

\vspace*{\myvspace}
\noindent \textbf{Fine-tuning vs. Training from Scratch} We assessed the impact of initializing our models with CroCo \cite{weinzaepfel2023croco} weights versus random initialization. 
Initializing with CroCo significantly improved performance in both the coarse estimation (rows \ref{ab.1} vs. \ref{ab.3}) and pose refinement stages (rows \ref{ab.4} vs. \ref{ab.6}).
Although the performance improvement in the pose selection stage (rows \ref{ab.7}, \ref{ab.8}) is relatively small than others, using CroCo pretraining enhances performance across all stages.

\vspace*{\myvspace}
\noindent \textbf{Hybrid Representation vs. Direct Regression} Comparing rows \ref{ab.1} and \ref{ab.2}, we evaluated the impact of our hybrid representation (combining patch-level classification and offset regression) against direct regression of correspondences.
Removing patch-level classification reduced the model’s generalization ability, demonstrating the effectiveness of our hybrid approach.

\noindent \textbf{Probabilistic Flow Regression} We evaluated the effectiveness of defining flow as a conditional probability by training a model with the same architecture as Co-op but using the loss function and prediction head from GenFlow \cite{moon2024genflow}.
The results (rows \ref{ab.4} vs. \ref{ab.5}) show that probabilistic flow regression improves flow learning and pose accuracy by estimating flow reliability, enabling the model to better handle ambiguous or unreliable correspondences.
\section{Conclusion}

We propose Co-op, a novel method for 6D pose estimation of unseen objects. Co-op can estimate the poses of novel objects using CAD models that are not provided during training but are available only during testing. We introduce a coarse estimation method that achieves strong generalization performance by leveraging a hybrid representation. Furthermore, we propose a render-and-compare refiner that employs probabilistic flow. Using these components, we demonstrate that our method significantly outperforms existing methods by a large margin on the BOP benchmark, which comprises hundreds of diverse objects and cluttered scenes.

\label{sec:conclusion}

{\small \noindent\textbf{Acknowledgments.} This work was supported by the Technology Innovation Program(00417108, Technologies of Rapid Digital Twin Development Tools for Robotic Service Environments) funded By the Ministry of Trade Industry \& Energy(MOTIE, Korea)}


{
    \small
    \bibliographystyle{ieeenat_fullname}
    \bibliography{main}

\begin{thebibliography}{74}
\providecommand{\natexlab}[1]{#1}
\providecommand{\url}[1]{\texttt{#1}}
\expandafter\ifx\csname urlstyle\endcsname\relax
  \providecommand{\doi}[1]{doi: #1}\else
  \providecommand{\doi}{doi: \begingroup \urlstyle{rm}\Url}\fi

\bibitem[pil()]{pillow}
The pillow imaging library. https://github.com/python-pillow/pillow.

\bibitem[Ausserlechner et~al.(2024)Ausserlechner, Haberger, Thalhammer, Weibel, and Vincze]{ausserlechner2024zs6d}
Philipp Ausserlechner, David Haberger, Stefan Thalhammer, Jean-Baptiste Weibel, and Markus Vincze.
\newblock Zs6d: Zero-shot 6d object pose estimation using vision transformers.
\newblock In \emph{ICRA}, pages 463--469. IEEE, 2024.

\bibitem[Barath et~al.(2020)Barath, Noskova, Ivashechkin, and Matas]{barath2019magsacplusplus}
Daniel Barath, Jana Noskova, Maksym Ivashechkin, and Jiri Matas.
\newblock {MAGSAC}++, a fast, reliable and accurate robust estimator.
\newblock In \emph{Conference on Computer Vision and Pattern Recognition}, 2020.

\bibitem[Bay et~al.(2006)Bay, Tuytelaars, and Van~Gool]{Bay2006surf}
Herbert Bay, Tinne Tuytelaars, and Luc Van~Gool.
\newblock Surf: Speeded up robust features.
\newblock In \emph{ECCV}, pages 404--417, Berlin, Heidelberg, 2006. Springer Berlin Heidelberg.

\bibitem[Brachmann et~al.(2014)Brachmann, Krull, Michel, Gumhold, Shotton, and Rother]{Brachmann2014LMO}
Eric Brachmann, Alexander Krull, Frank Michel, Stefan Gumhold, Jamie Shotton, and Carsten Rother.
\newblock Learning 6d object pose estimation using 3d object coordinates.
\newblock In \emph{ECCV}, 2014.

\bibitem[Calonder et~al.(2012)Calonder, Lepetit, Ozuysal, Trzcinski, Strecha, and Fua]{Calonder2012brief}
Michael Calonder, Vincent Lepetit, Mustafa Ozuysal, Tomasz Trzcinski, Christoph Strecha, and Pascal Fua.
\newblock Brief: Computing a local binary descriptor very fast.
\newblock \emph{IEEE Transactions on Pattern Analysis and Machine Intelligence}, 34\penalty0 (7):\penalty0 1281--1298, 2012.

\bibitem[Caraffa et~al.(2024)Caraffa, Boscaini, Hamza, and Poiesi]{caraffa2024freeze}
Andrea Caraffa, Davide Boscaini, Amir Hamza, and Fabio Poiesi.
\newblock Freeze: Training-free zero-shot 6d pose estimation with geometric and vision foundation models.
\newblock In \emph{ECCV}, 2024.

\bibitem[Chang et~al.(2015)Chang, Funkhouser, Guibas, Hanrahan, Huang, Li, Savarese, Savva, Song, Su, et~al.]{chang2015shapenet}
Angel~X Chang, Thomas Funkhouser, Leonidas Guibas, Pat Hanrahan, Qixing Huang, Zimo Li, Silvio Savarese, Manolis Savva, Shuran Song, Hao Su, et~al.
\newblock Shapenet: An information-rich 3d model repository.
\newblock \emph{arXiv preprint arXiv:1512.03012}, 2015.

\bibitem[Chen et~al.(2022{\natexlab{a}})Chen, Luo, Zhou, Tian, Zhen, Fang, McKinnon, Tsin, and Quan]{chen2022aspanformer}
Hongkai Chen, Zixin Luo, Lei Zhou, Yurun Tian, Mingmin Zhen, Tian Fang, David McKinnon, Yanghai Tsin, and Long Quan.
\newblock Aspanformer: Detector-free image matching with adaptive span transformer.
\newblock In \emph{ECCV}, 2022{\natexlab{a}}.

\bibitem[Chen et~al.(2022{\natexlab{b}})Chen, Wang, Wang, Tian, Xiong, and Li]{chen2022epro}
Hansheng Chen, Pichao Wang, Fan Wang, Wei Tian, Lu Xiong, and Hao Li.
\newblock Epro-pnp: Generalized end-to-end probabilistic perspective-n-points for monocular object pose estimation.
\newblock In \emph{CVPR}, pages 2781--2790, 2022{\natexlab{b}}.

\bibitem[Denninger et~al.(2023)Denninger, Winkelbauer, Sundermeyer, Boerdijk, Knauer, Strobl, Humt, and Triebel]{Denninger2023}
Maximilian Denninger, Dominik Winkelbauer, Martin Sundermeyer, Wout Boerdijk, Markus Knauer, Klaus~H. Strobl, Matthias Humt, and Rudolph Triebel.
\newblock Blenderproc2: A procedural pipeline for photorealistic rendering.
\newblock \emph{Journal of Open Source Software}, 8\penalty0 (82):\penalty0 4901, 2023.

\bibitem[DeTone et~al.(2018)DeTone, Malisiewicz, and Rabinovich]{DeTone2018superpoint}
Daniel DeTone, Tomasz Malisiewicz, and Andrew Rabinovich.
\newblock Superpoint: Self-supervised interest point detection and description.
\newblock In \emph{CVPRW}, pages 337--33712, 2018.

\bibitem[Dosovitskiy et~al.(2021)Dosovitskiy, Beyer, Kolesnikov, Weissenborn, Zhai, Unterthiner, Dehghani, Minderer, Heigold, Gelly, Uszkoreit, and Houlsby]{dosovitskiy2020vit}
Alexey Dosovitskiy, Lucas Beyer, Alexander Kolesnikov, Dirk Weissenborn, Xiaohua Zhai, Thomas Unterthiner, Mostafa Dehghani, Matthias Minderer, Georg Heigold, Sylvain Gelly, Jakob Uszkoreit, and Neil Houlsby.
\newblock An image is worth 16x16 words: Transformers for image recognition at scale.
\newblock In \emph{ICLR}, 2021.

\bibitem[Doumanoglou et~al.(2016)Doumanoglou, Kouskouridas, Malassiotis, and Kim]{DoumanoglouKMK16ICBIN}
Andreas Doumanoglou, Rigas Kouskouridas, Sotiris Malassiotis, and Tae-Kyun Kim.
\newblock Recovering 6d object pose and predicting next-best-view in the crowd.
\newblock In \emph{CVPR}, pages 3583--3592, 2016.

\bibitem[Downs et~al.(2022)Downs, Francis, Koenig, Kinman, Hickman, Reymann, McHugh, and Vanhoucke]{downs2022google}
Laura Downs, Anthony Francis, Nate Koenig, Brandon Kinman, Ryan Hickman, Krista Reymann, Thomas~B McHugh, and Vincent Vanhoucke.
\newblock Google scanned objects: A high-quality dataset of 3d scanned household items.
\newblock In \emph{ICRA}, pages 2553--2560. IEEE, 2022.

\bibitem[Drost et~al.(2017)Drost, Ulrich, Bergmann, Härtinger, and Steger]{DrostUBHS17ITODD}
Bertram Drost, Markus Ulrich, Paul Bergmann, Philipp Härtinger, and Carsten Steger.
\newblock Introducing mvtec itodd — a dataset for 3d object recognition in industry.
\newblock In \emph{ICCVW}, pages 2200--2208, 2017.

\bibitem[Dusmanu et~al.(2019)Dusmanu, Rocco, Pajdla, Pollefeys, Sivic, Torii, and Sattler]{Dusmanu2019CVPR}
Mihai Dusmanu, Ignacio Rocco, Tomas Pajdla, Marc Pollefeys, Josef Sivic, Akihiko Torii, and Torsten Sattler.
\newblock {D2-Net: A Trainable CNN for Joint Detection and Description of Local Features}.
\newblock In \emph{CVPR}, 2019.

\bibitem[Edstedt et~al.(2023)Edstedt, Athanasiadis, Wadenbäck, and Felsberg]{edstedt2023dkm}
Johan Edstedt, Ioannis Athanasiadis, Mårten Wadenbäck, and Michael Felsberg.
\newblock {DKM}: Dense kernelized feature matching for geometry estimation.
\newblock In \emph{CVPR}, 2023.

\bibitem[Edstedt et~al.(2024)Edstedt, Sun, Bökman, Wadenbäck, and Felsberg]{edstedt2024roma}
Johan Edstedt, Qiyu Sun, Georg Bökman, Mårten Wadenbäck, and Michael Felsberg.
\newblock {RoMa: Robust Dense Feature Matching}.
\newblock In \emph{CVPR}, 2024.

\bibitem[Fischler and Bolles(1981)]{FischlerB81RANSAC}
Martin~A. Fischler and Robert~C. Bolles.
\newblock Random sample consensus: {A} paradigm for model fitting with applications to image analysis and automated cartography.
\newblock \emph{Commun. {ACM}}, 24\penalty0 (6):\penalty0 381--395, 1981.

\bibitem[Goslin and Mine(2004)]{1350741}
M. Goslin and M.R. Mine.
\newblock The panda3d graphics engine.
\newblock \emph{Computer}, 37\penalty0 (10):\penalty0 112--114, 2004.

\bibitem[Haugaard and Buch(2022)]{haugaard2022surfemb}
Rasmus~Laurvig Haugaard and Anders~Glent Buch.
\newblock Surfemb: Dense and continuous correspondence distributions for object pose estimation with learnt surface embeddings.
\newblock In \emph{CVPR}, pages 6749--6758, 2022.

\bibitem[He et~al.(2022)He, Sun, Wang, Huang, Bao, and Zhou]{he2022oneposeplusplus}
Xingyi He, Jiaming Sun, Yuang Wang, Di Huang, Hujun Bao, and Xiaowei Zhou.
\newblock Onepose++: Keypoint-free one-shot object pose estimation without {CAD} models.
\newblock In \emph{NeurIPS}, 2022.

\bibitem[Hoda{\v{n}} et~al.(2017)Hoda{\v{n}}, Haluza, Obdr{\v{z}}{\'a}lek, Matas, Lourakis, and Zabulis]{hodan2017tless}
Tom{\'a}{\v{s}} Hoda{\v{n}}, Pavel Haluza, {\v{S}}t{\v{e}}p{\'a}n Obdr{\v{z}}{\'a}lek, Ji{\v{r}}{\'\i} Matas, Manolis Lourakis, and Xenophon Zabulis.
\newblock {T-LESS}: An {RGB-D} dataset for {6D} pose estimation of texture-less objects.
\newblock \emph{WACV}, 2017.

\bibitem[Hoda{\v{n}} et~al.(2018)Hoda{\v{n}}, Michel, Brachmann, Kehl, Glent~Buch, Kraft, Drost, Vidal, Ihrke, Zabulis, Sahin, Manhardt, Tombari, Kim, Matas, and Rother]{hodan2018TUDL}
Tom{\'a}{\v{s}} Hoda{\v{n}}, Frank Michel, Eric Brachmann, Wadim Kehl, Anders Glent~Buch, Dirk Kraft, Bertram Drost, Joel Vidal, Stephan Ihrke, Xenophon Zabulis, Caner Sahin, Fabian Manhardt, Federico Tombari, Tae-Kyun Kim, Ji{\v{r}}{\'i} Matas, and Carsten Rother.
\newblock {BOP}: Benchmark for {6D} object pose estimation.
\newblock \emph{ECCV}, 2018.

\bibitem[Hoda{\v{n}} et~al.(2020)Hoda{\v{n}}, Sundermeyer, Drost, Labb{\'e}, Brachmann, Michel, Rother, and Matas]{hodavn2020bop}
Tom{\'a}{\v{s}} Hoda{\v{n}}, Martin Sundermeyer, Bertram Drost, Yann Labb{\'e}, Eric Brachmann, Frank Michel, Carsten Rother, and Ji{\v{r}}{\'\i} Matas.
\newblock Bop challenge 2020 on 6d object localization.
\newblock In \emph{Computer Vision--ECCV 2020 Workshops: Glasgow, UK, August 23--28, 2020, Proceedings, Part II 16}, pages 577--594. Springer, 2020.

\bibitem[Hodan et~al.(2024)Hodan, Sundermeyer, Labbe, Nguyen, Wang, Brachmann, Drost, Lepetit, Rother, and Matas]{hodan2024bop}
Tomas Hodan, Martin Sundermeyer, Yann Labbe, Van~Nguyen Nguyen, Gu Wang, Eric Brachmann, Bertram Drost, Vincent Lepetit, Carsten Rother, and Jiri Matas.
\newblock Bop challenge 2023 on detection segmentation and pose estimation of seen and unseen rigid objects.
\newblock In \emph{Proceedings of the IEEE/CVF Conference on Computer Vision and Pattern Recognition}, pages 5610--5619, 2024.

\bibitem[Hu et~al.(2022)Hu, Fua, and Salzmann]{hu2022perspective}
Yinlin Hu, Pascal Fua, and Mathieu Salzmann.
\newblock Perspective flow aggregation for data-limited 6d object pose estimation.
\newblock In \emph{ECCV}, pages 89--106. Springer, 2022.

\bibitem[Kabsch(1976)]{kabsch1976solution}
Wolfgang Kabsch.
\newblock A solution for the best rotation to relate two sets of vectors.
\newblock \emph{Acta Crystallographica Section A: Crystal Physics, Diffraction, Theoretical and General Crystallography}, 32\penalty0 (5):\penalty0 922--923, 1976.

\bibitem[Kaskman et~al.(2019)Kaskman, Zakharov, Shugurov, and Ilic]{KaskmanZSI19HB}
Roman Kaskman, Sergey Zakharov, Ivan Shugurov, and Slobodan Ilic.
\newblock Homebreweddb: Rgb-d dataset for 6d pose estimation of 3d objects.
\newblock In \emph{ICCVW}, pages 2767--2776, 2019.

\bibitem[Kehl et~al.(2017)Kehl, Manhardt, Tombari, Ilic, and Navab]{kehl2017ssd}
Wadim Kehl, Fabian Manhardt, Federico Tombari, Slobodan Ilic, and Nassir Navab.
\newblock Ssd-6d: Making rgb-based 3d detection and 6d pose estimation great again.
\newblock In \emph{ICCV}, pages 1521--1529, 2017.

\bibitem[Kirillov et~al.(2023)Kirillov, Mintun, Ravi, Mao, Rolland, Gustafson, Xiao, Whitehead, Berg, Lo, et~al.]{kirillov2023segment}
Alexander Kirillov, Eric Mintun, Nikhila Ravi, Hanzi Mao, Chloe Rolland, Laura Gustafson, Tete Xiao, Spencer Whitehead, Alexander~C Berg, Wan-Yen Lo, et~al.
\newblock Segment anything.
\newblock In \emph{ICCV}, pages 4015--4026, 2023.

\bibitem[{Labbe} et~al.(2020){Labbe}, {Carpentier}, {Aubry}, and {Sivic}]{labbe2020}
Y. {Labbe}, J. {Carpentier}, M. {Aubry}, and J. {Sivic}.
\newblock Cosypose: Consistent multi-view multi-object 6d pose estimation.
\newblock In \emph{ECCV}, 2020.

\bibitem[Labb\'e et~al.(2022)Labb\'e, Manuelli, Mousavian, Tyree, Birchfield, Tremblay, Carpentier, Aubry, Fox, and Sivic]{labbe2022megapose}
Yann Labb\'e, Lucas Manuelli, Arsalan Mousavian, Stephen Tyree, Stan Birchfield, Jonathan Tremblay, Justin Carpentier, Mathieu Aubry, Dieter Fox, and Josef Sivic.
\newblock {{MegaPose}}: {{6D Pose Estimation}} of {{Novel Objects}} via {{Render}} \& {{Compare}}.
\newblock In \emph{CoRL}, 2022.

\bibitem[Lepetit et~al.(2009)Lepetit, Moreno-Noguer, and Fua]{Lepetit2009ep}
Vincent Lepetit, Francesc Moreno-Noguer, and Pascal Fua.
\newblock Epnp: An accurate o(n) solution to the pnp problem.
\newblock \emph{International Journal Of Computer Vision}, 81:\penalty0 155--166, 2009.

\bibitem[Li et~al.(2023)Li, Vutukur, Yu, Shugurov, Busam, Yang, and Ilic]{li2023nerf}
Fu Li, Shishir~Reddy Vutukur, Hao Yu, Ivan Shugurov, Benjamin Busam, Shaowu Yang, and Slobodan Ilic.
\newblock Nerf-pose: A first-reconstruct-then-regress approach for weakly-supervised 6d object pose estimation.
\newblock In \emph{ICCV}, pages 2123--2133, 2023.

\bibitem[Li et~al.(2018)Li, Wang, Ji, Xiang, and Fox]{li2017deepim}
Yi Li, Gu Wang, Xiangyang Ji, Yu Xiang, and Dieter Fox.
\newblock {DeepIM}: Deep iterative matching for {6D} pose estimation.
\newblock In \emph{ECCV}, 2018.

\bibitem[Li et~al.(2019)Li, Wang, and Ji]{li2019cdpn}
Zhigang Li, Gu Wang, and Xiangyang Ji.
\newblock Cdpn: Coordinates-based disentangled pose network for real-time rgb-based 6-dof object pose estimation.
\newblock In \emph{ICCV}, pages 7678--7687, 2019.

\bibitem[Lin et~al.(2024)Lin, Liu, Lu, and Jia]{Lin2024sam6d}
Jiehong Lin, Lihua Liu, Dekun Lu, and Kui Jia.
\newblock Sam-6d: Segment anything model meets zero-shot 6d object pose estimation.
\newblock In \emph{CVPR}, pages 27906--27916, 2024.

\bibitem[Lin et~al.(2022)Lin, Tremblay, Tyree, Vela, and Birchfield]{lin2022icra:centerpose}
Yunzhi Lin, Jonathan Tremblay, Stephen Tyree, Patricio~A. Vela, and Stan Birchfield.
\newblock Single-stage keypoint-based category-level object pose estimation from an {RGB} image.
\newblock In \emph{ICRA}, 2022.

\bibitem[Lipson et~al.(2022)Lipson, Teed, Goyal, and Deng]{lipson2022coupled}
Lahav Lipson, Zachary Teed, Ankit Goyal, and Jia Deng.
\newblock Coupled iterative refinement for 6d multi-object pose estimation.
\newblock In \emph{CVPR}, pages 6728--6737, 2022.

\bibitem[Liu et~al.(2022{\natexlab{a}})Liu, Zhang, Zhang, Fu, Tang, Liang, Tang, Cheng, Zhang, Wang, and Ji]{liu2022gdrnpp_bop}
Xingyu Liu, Ruida Zhang, Chenyangguang Zhang, Bowen Fu, Jiwen Tang, Xiquan Liang, Jingyi Tang, Xiaotian Cheng, Yukang Zhang, Gu Wang, and Xiangyang Ji.
\newblock Gdrnpp.
\newblock \url{https://github.com/shanice-l/gdrnpp_bop2022}, 2022{\natexlab{a}}.

\bibitem[Liu et~al.(2022{\natexlab{b}})Liu, Wen, Peng, Lin, Long, Komura, and Wang]{liu2022gen6d}
Yuan Liu, Yilin Wen, Sida Peng, Cheng Lin, Xiaoxiao Long, Taku Komura, and Wenping Wang.
\newblock Gen6d: Generalizable model-free 6-dof object pose estimation from rgb images.
\newblock In \emph{ECCV}, 2022{\natexlab{b}}.

\bibitem[Liu et~al.(2022{\natexlab{c}})Liu, Mao, Wu, Feichtenhofer, Darrell, and Xie]{liu2022convnet}
Zhuang Liu, Hanzi Mao, Chao-Yuan Wu, Christoph Feichtenhofer, Trevor Darrell, and Saining Xie.
\newblock A convnet for the 2020s.
\newblock In \emph{CVPR}, pages 11976--11986, 2022{\natexlab{c}}.

\bibitem[Loshchilov and Hutter(2017)]{Loshchilov2017DecoupledWD}
Ilya Loshchilov and Frank Hutter.
\newblock Decoupled weight decay regularization.
\newblock In \emph{ICLR}, 2017.

\bibitem[Lowe(2004)]{Lowe2004DistinctiveIF}
David~G. Lowe.
\newblock Distinctive image features from scale-invariant keypoints.
\newblock \emph{International Journal of Computer Vision}, 60:\penalty0 91--110, 2004.

\bibitem[Moon et~al.(2024)Moon, Son, Hur, and Kim]{moon2024genflow}
Sungphill Moon, Hyeontae Son, Dongcheol Hur, and Sangwook Kim.
\newblock Genflow: Generalizable recurrent flow for 6d pose refinement of novel objects.
\newblock In \emph{CVPR}, pages 10039--10049, 2024.

\bibitem[Nguyen et~al.(2022)Nguyen, Hu, Xiao, Salzmann, and Lepetit]{Nguyen2022template}
Van~Nguyen Nguyen, Yinlin Hu, Yang Xiao, Mathieu Salzmann, and Vincent Lepetit.
\newblock Templates for 3d object pose estimation revisited: Generalization to new objects and robustness to occlusions.
\newblock In \emph{CVPR}, pages 6771--6780, 2022.

\bibitem[Nguyen et~al.(2023)Nguyen, Groueix, Ponimatkin, Lepetit, and Hodan]{nguyen2023cnos}
Van~Nguyen Nguyen, Thibault Groueix, Georgy Ponimatkin, Vincent Lepetit, and Tomas Hodan.
\newblock Cnos: A strong baseline for cad-based novel object segmentation.
\newblock In \emph{ICCVW}, pages 2134--2140, 2023.

\bibitem[Nguyen et~al.(2024)Nguyen, Groueix, Salzmann, and Lepetit]{nguyen2024gigapose}
Van~Nguyen Nguyen, Thibault Groueix, Mathieu Salzmann, and Vincent Lepetit.
\newblock Gigapose: Fast and robust novel object pose estimation via one correspondence.
\newblock In \emph{CVPR}, pages 9903--9913, 2024.

\bibitem[Okorn et~al.(2021)Okorn, Gu, Hebert, and Held]{okorn2021zephyr}
Brian Okorn, Qiao Gu, Martial Hebert, and David Held.
\newblock Zephyr: Zero-shot pose hypothesis rating.
\newblock In \emph{ICRA}, pages 14141--14148. IEEE, 2021.

\bibitem[Oquab et~al.(2023)Oquab, Darcet, Moutakanni, Vo, Szafraniec, Khalidov, Fernandez, Haziza, Massa, El-Nouby, et~al.]{oquab2023dinov2}
Maxime Oquab, Timoth{\'e}e Darcet, Th{\'e}o Moutakanni, Huy Vo, Marc Szafraniec, Vasil Khalidov, Pierre Fernandez, Daniel Haziza, Francisco Massa, Alaaeldin El-Nouby, et~al.
\newblock Dinov2: Learning robust visual features without supervision.
\newblock \emph{arXiv preprint arXiv:2304.07193}, 2023.

\bibitem[{\"O}rnek et~al.(2024){\"O}rnek, Labb\'e, Tekin, Ma, Keskin, Forster, and Hoda{\v{n}}]{ornek2024foundpose}
Evin~P{\i}nar {\"O}rnek, Yann Labb\'e, Bugra Tekin, Lingni Ma, Cem Keskin, Christian Forster, and Tom{\'a}{\v{s}} Hoda{\v{n}}.
\newblock Foundpose: Unseen object pose estimation with foundation features.
\newblock In \emph{ECCV}, 2024.

\bibitem[Park et~al.(2019)Park, Patten, and Vincze]{park2019pix2pose}
Kiru Park, Timothy Patten, and Markus Vincze.
\newblock Pix2pose: Pixel-wise coordinate regression of objects for 6d pose estimation.
\newblock In \emph{ICCV}, pages 7668--7677, 2019.

\bibitem[Paszke et~al.(2019)Paszke, Gross, Massa, Lerer, Bradbury, Chanan, Killeen, Lin, Gimelshein, Antiga, et~al.]{paszke2019pytorch}
Adam Paszke, Sam Gross, Francisco Massa, Adam Lerer, James Bradbury, Gregory Chanan, Trevor Killeen, Zeming Lin, Natalia Gimelshein, Luca Antiga, et~al.
\newblock Pytorch: An imperative style, high-performance deep learning library.
\newblock \emph{Advances in neural information processing systems}, 32, 2019.

\bibitem[Peng et~al.(2019)Peng, Liu, Huang, Zhou, and Bao]{peng2019pvnet}
Sida Peng, Yuan Liu, Qixing Huang, Xiaowei Zhou, and Hujun Bao.
\newblock Pvnet: Pixel-wise voting network for 6dof pose estimation.
\newblock In \emph{CVPR}, 2019.

\bibitem[Rad and Lepetit(2017)]{rad2017bb8}
Mahdi Rad and Vincent Lepetit.
\newblock Bb8: A scalable, accurate, robust to partial occlusion method for predicting the 3d poses of challenging objects without using depth.
\newblock In \emph{ICCV}, 2017.

\bibitem[Ranftl et~al.(2021)Ranftl, Bochkovskiy, and Koltun]{ranftl2021vision}
Ren{\'e} Ranftl, Alexey Bochkovskiy, and Vladlen Koltun.
\newblock Vision transformers for dense prediction.
\newblock In \emph{ICCV}, pages 12179--12188, 2021.

\bibitem[Revaud et~al.(2019)Revaud, Weinzaepfel, de~Souza, and Humenberger]{r2d2}
Jerome Revaud, Philippe Weinzaepfel, C{\'{e}}sar~Roberto de Souza, and Martin Humenberger.
\newblock {R2D2:} repeatable and reliable detector and descriptor.
\newblock In \emph{NeurIPS}, 2019.

\bibitem[Rublee et~al.(2011)Rublee, Rabaud, Konolige, and Bradski]{Rublee2012orb}
Ethan Rublee, Vincent Rabaud, Kurt Konolige, and Gary Bradski.
\newblock Orb: An efficient alternative to sift or surf.
\newblock In \emph{ICCV}, pages 2564--2571, 2011.

\bibitem[Shugurov et~al.(2022)Shugurov, Li, Busam, and Ilic]{shugurov2022osop}
Ivan Shugurov, Fu Li, Benjamin Busam, and Slobodan Ilic.
\newblock Osop: A multi-stage one shot object pose estimation framework.
\newblock In \emph{CVPR}, pages 6835--6844, 2022.

\bibitem[Sun et~al.(2021)Sun, Shen, Wang, Bao, and Zhou]{sun2021loftr}
Jiaming Sun, Zehong Shen, Yuang Wang, Hujun Bao, and Xiaowei Zhou.
\newblock {LoFTR}: Detector-free local feature matching with transformers.
\newblock In \emph{CVPR}, 2021.

\bibitem[Sun et~al.(2022)Sun, Wang, Zhang, He, Zhao, Zhang, and Zhou]{sun2022onepose}
Jiaming Sun, Zihao Wang, Siyu Zhang, Xingyi He, Hongcheng Zhao, Guofeng Zhang, and Xiaowei Zhou.
\newblock {OnePose}: One-shot object pose estimation without {CAD} models.
\newblock In \emph{CVPR}, 2022.

\bibitem[Tekin et~al.(2018)Tekin, Sinha, and Fua]{tekin18}
Bugra Tekin, Sudipta~N. Sinha, and Pascal Fua.
\newblock {Real-Time Seamless Single Shot 6D Object Pose Prediction}.
\newblock In \emph{CVPR}, 2018.

\bibitem[Truong et~al.(2020)Truong, Danelljan, and Timofte]{truong2020glu}
Prune Truong, Martin Danelljan, and Radu Timofte.
\newblock Glu-net: Global-local universal network for dense flow and correspondences.
\newblock In \emph{CVPR}, pages 6258--6268, 2020.

\bibitem[Truong et~al.(2023)Truong, Danelljan, Timofte, and Van~Gool]{truong2023pdc}
Prune Truong, Martin Danelljan, Radu Timofte, and Luc Van~Gool.
\newblock Pdc-net+: Enhanced probabilistic dense correspondence network.
\newblock \emph{IEEE Transactions on Pattern Analysis and Machine Intelligence}, 45\penalty0 (8):\penalty0 10247--10266, 2023.

\bibitem[Tyszkiewicz et~al.(2020)Tyszkiewicz, Fua, and Trulls]{tyszkiewicz2020disk}
Micha{\l} Tyszkiewicz, Pascal Fua, and Eduard Trulls.
\newblock Disk: Learning local features with policy gradient.
\newblock In \emph{NeurIPS}, 2020.

\bibitem[Wang et~al.(2021)Wang, Manhardt, Tombari, and Ji]{Wang_2021_GDRN}
Gu Wang, Fabian Manhardt, Federico Tombari, and Xiangyang Ji.
\newblock {GDR-Net}: Geometry-guided direct regression network for monocular 6d object pose estimation.
\newblock In \emph{CVPR}, pages 16611--16621, 2021.

\bibitem[Wang et~al.(2019)Wang, Sridhar, Huang, Valentin, Song, and Guibas]{Wang_2019_CVPR}
He Wang, Srinath Sridhar, Jingwei Huang, Julien Valentin, Shuran Song, and Leonidas~J. Guibas.
\newblock Normalized object coordinate space for category-level 6d object pose and size estimation.
\newblock In \emph{CVPR}, 2019.

\bibitem[Wang et~al.(2024)Wang, He, Peng, Tan, and Zhou]{wang2024eloftr}
Yifan Wang, Xingyi He, Sida Peng, Dongli Tan, and Xiaowei Zhou.
\newblock {Efficient LoFTR}: Semi-dense local feature matching with sparse-like speed.
\newblock In \emph{CVPR}, 2024.

\bibitem[Weinzaepfel et~al.(2023)Weinzaepfel, Lucas, Leroy, Cabon, Arora, Br{\'e}gier, Csurka, Antsfeld, Chidlovskii, and Revaud]{weinzaepfel2023croco}
Philippe Weinzaepfel, Thomas Lucas, Vincent Leroy, Yohann Cabon, Vaibhav Arora, Romain Br{\'e}gier, Gabriela Csurka, Leonid Antsfeld, Boris Chidlovskii, and J{\'e}r{\^o}me Revaud.
\newblock Croco v2: Improved cross-view completion pre-training for stereo matching and optical flow.
\newblock In \emph{ICCV}, pages 17969--17980, 2023.

\bibitem[Wen et~al.(2024)Wen, Yang, Kautz, and Birchfield]{wen2024foundationpose}
Bowen Wen, Wei Yang, Jan Kautz, and Stan Birchfield.
\newblock Foundationpose: Unified 6d pose estimation and tracking of novel objects.
\newblock In \emph{CVPR}, pages 17868--17879, 2024.

\bibitem[Xiang et~al.(2018)Xiang, Schmidt, Narayanan, and Fox]{xiang2018posecnn}
Yu Xiang, Tanner Schmidt, Venkatraman Narayanan, and Dieter Fox.
\newblock Posecnn: A convolutional neural network for 6d object pose estimation in cluttered scenes.
\newblock In \emph{RSS}, 2018.

\bibitem[Yang et~al.(2020)Yang, Shi, and Carlone]{Yang20tro-teaser}
H. Yang, J. Shi, and L. Carlone.
\newblock {TEASER: Fast and Certifiable Point Cloud Registration}.
\newblock \emph{{IEEE} Trans. Robotics}, 2020.

\end{thebibliography}
}

\clearpage
\setcounter{page}{1}
\maketitlesupplementary

\section{Training Details}
\label{sec:training_detail}

In this section, we provid the model configurations, learning rates, and training schedules used for training each model.
All three models of Co-op are based on CroCo \cite{weinzaepfel2023croco}.
Each model takes two images as input and extracts features using a ViT \cite{dosovitskiy2020vit} encoder with shared weights.
The ViT decoder then processes these two sets of features together and estimates the required outputs through each model’s respective head.
We implemented our method using PyTorch \cite{paszke2019pytorch} and the Panda3D renderer \cite{1350741}.
All models are trained using eight NVIDIA A100 GPUs, and we used AdamW \cite{Loshchilov2017DecoupledWD} as the optimizer.

\subsection{Coarse Model}

For the coarse model, we use ViT-Large for the encoder and ViT-Base for the decoder.
We train the model with a batch size of 64 over 450 epochs, where each epoch consists of 1,800 iterations.
The learning rate decreases from $2.0\times 10^{-5}$ to $2.0\times 10^{-6}$ after 250 epochs, with a warm-up period during the first 50 epochs.
Training the coarse model takes approximately two days.

\subsection{Refiner Model}

The refiner model uses the same encoder and decoder architecture as the coarse model.
During training, the batch size is 32, and we use the same learning rate and training schedule as the coarse model.
Unlike the coarse model, we use gradient clipping to prevent excessively large gradients caused by unstable correspondences during the early training stages.
The gradient clipping value is set to $10^{-2}$.
Training the refiner model takes approximately three days.

\noindent \textbf{Pose Loss} To train the refiner model, we need a pose loss \(\mathcal{L}_{pose}\) in addition to the \(\mathcal{L}_{cert}\) and \(\mathcal{L}_{flow}\) losses described in the main paper. We define \(\mathcal{L}_{pose}\) as a disentangled point matching loss, the same as in GenFlow \cite{moon2024genflow}.

Given the estimated pose $\mathbf{P} = [\mathbf{R} \mid [\mathbf{t}_x, \mathbf{t}_y, \mathbf{t}_z]^T]$ and the ground truth pose $\mathbf{P}_{gt} = [\bar{\mathbf{R}} \mid [\bar{\mathbf{t}}_x, \bar{\mathbf{t}}_y, \bar{\mathbf{t}}_z]^T]$, $\mathcal{L}_{pose}$ is defined as follows:

\begin{align}
	\begin{split}
		\mathcal{L}_{pose} = \mathcal{D}([\mathbf{R}|[\bar{\mathbf{t}}_x, \bar{\mathbf{t}}_y, \bar{\mathbf{t}}_z]^{\text{T}}], \mathbf{P}_{gt})  \\
		+ \mathcal{D}([\bar{\mathbf{R}}|[\mathbf{t}_{x}, \mathbf{t}_{y}, \bar{\mathbf{t}}_z]^{\text{T}}], \mathbf{P}_{gt}) \\
		+ \mathcal{D}([\bar{\mathbf{R}}|[\bar{\mathbf{t}}_x, \bar{\mathbf{t}}_y, \mathbf{t}_{z}]^{\text{T}}], \mathbf{P}_{gt}).
	\end{split}
\end{align}

Here, $\mathcal{D}$ is the average L1 distance between two sets of 3D points obtained by transforming the 3D points on the object’s surface using the poses $\mathbf{P}$ and $\mathbf{P}_{gt}$, respectively.

\subsection{Selection Model}

The selection model uses a smaller encoder and decoder architecture compared to the coarse and refiner models.
Specifically, we use ViT-Base for the encoder and ViT-Small for the decoder.
We train the model for 200 epochs with a batch size of 16, where each epoch consists of 7,200 iterations.
The learning rate is $2.0 \times 10^{-5}$, and similar to the other models, we have a warm-up period during the first 50 epochs to stabilize the training process.
Training the selection model takes approximately 1.5 days.

\subsection{Data Augmentation}
During model training, we apply random perturbations to the training RGB images to enhance robustness against domain shifts.
We employ the same data augmentation techniques as MegaPose \cite{labbe2022megapose}, which include Gaussian blur, contrast adjustment, brightness adjustment, and color filtering using the Pillow library \cite{pillow}.

\setlength{\tabcolsep}{2.5pt}
\begin{table*}[t!]
    \begin{center}
    \scriptsize
    \begin{tabularx}{1.0\linewidth}{r l |>{\centering\arraybackslash}p{2.8cm}|Y Y Y Y Y Y Y |Y| Y}
    \toprule
       \# \vspace{0.4ex} &
       Method &
       Detection / Segmentation &
       LM-O &
       T-LESS &
       TUD-L &
       IC-BIN &
       ITODD &
       HB &
       YCB-V &
       Mean &
       Time(sec)\\
    \toprule

    \newtag{1}{supp.1} & MegaPose \cite{labbe2022megapose} & \multirow{8}{*}{CNOS \cite{nguyen2023cnos}}  & 62.6 & 48.7 & 85.1 & 46.7 & 46.8 & 73.0 & 76.4 & 62.8 & 141.965 \\

    \newtag{2}{supp.2} & SAM-6D \cite{Lin2024sam6d} &  & 65.1 & 47.9 & 82.5 & 49.7 & 56.2 & 73.8 & 81.5 & 65.3 & \phantom{0}1.254 \\

    \newtag{3}{supp.3} & GenFlow \cite{moon2024genflow} &  & 63.5 & 52.1 & 86.2 & 53.4 & 55.4 & 77.9 & 83.3 & 67.4 & 34.578 \\

    \newtag{4}{supp.4} & GigaPose \cite{nguyen2024gigapose} + GenFlow \cite{moon2024genflow} &  & 67.8 & 55.6 & 81.1 & 56.3 & 57.5 & 79.1 & 82.5 & 68.6 & 11.140 \\ 

    \newtag{5}{supp.5} & FreeZe \cite{caraffa2024freeze} &  & 68.9	& 52.0 & \textbf{93.6} & 49.9 & 56.1	& 79.0 & 85.3 & 69.3 & 13.474 \\

    \rowcolor[gray]{0.9} \newtag{6}{supp.6} & Co-op (Coarse) &   & 70.0 & 64.2 &  87.9 & 56.4 & 56.6 & 84.2 & 85.3 & 72.1 & \phantom{0}0.978 \\

    \rowcolor[gray]{0.9} \newtag{7}{supp.7} & Co-op (1 Hypotheses) &  & \underline{71.5} & \underline{64.6} & \underline{90.5} & \underline{57.5} & \underline{58.2} & \underline{85.7} & \underline{87.4} & \underline{73.6} & \phantom{0}2.331 \\

    \rowcolor[gray]{0.9} \newtag{8}{supp.8} & Co-op (5 Hypotheses) &  & \textbf{73.0} & \textbf{66.4} & \underline{90.5} & \textbf{59.7} & \textbf{61.3} & \textbf{87.1} & \textbf{88.7} & \textbf{75.2} & \phantom{0}7.162 \\
    
    \midrule

    \newtag{9}{supp.9} & SAM-6D \cite{Lin2024sam6d} & \multirow{6}{*}{SAM-6D \cite{Lin2024sam6d}} & 69.9 & 51.5 & 90.4 & \underline{58.8} & \underline{60.2} & 77.6 & 84.5 & 70.4 & \phantom{0}4.367 \\

    \newtag{10}{supp.10} & FreeZe \cite{caraffa2024freeze} & & 71.6 & 53.1 & \textbf{94.9} & 54.5 & 58.6 & 79.6 & 84.0 & 70.9 & 11.473 \\

    \newtag{11}{supp.11} & FoundationPose \cite{wen2024foundationpose} & & \textbf{75.6} & 64.6 & \underline{92.3} & 50.8 & 58.0 & 83.5 & \textbf{88.9} & 73.4 & 29.317 \\

    \rowcolor[gray]{0.9} \newtag{12}{supp.12} & Co-op (Coarse) & \multirow{-2}{*}{SAM-6D \cite{Lin2024sam6d}} & 70.0 & 65.4 & 89.2 & 56.6 & 58.4 & 83.7 & 85.5 & 72.7 & \phantom{0}0.911  \\

    \rowcolor[gray]{0.9} \newtag{13}{supp.13} & Co-op (1 Hypotheses) & & 71.6 & \underline{65.8} & 91.5 & 57.6 & 60.1 & \underline{85.1} & \underline{87.7} & \underline{74.2} & \phantom{0}2.265  \\

    \rowcolor[gray]{0.9} \newtag{14}{supp.14} & Co-op (5 Hypotheses) & & \underline{73.0} & \textbf{67.8} & 91.5 & \textbf{59.8} & \textbf{63.1} & \textbf{87.5} & \textbf{88.9} & \textbf{75.9} & \phantom{0}7.236  \\
    
    \bottomrule
    \end{tabularx}
    \end{center}
    \caption{\textbf{BOP Benchmark Results Using RGB-D} Results on the seven core datasets of the BOP challenge using RGB-D inputs.
    }
    \label{tab:supp_results}
\end{table*}

\section{Additional Experiments}

\subsection{BOP Benchmark Results Using RGB-D}

Although Co-op was trained for object pose estimation using single RGB image inputs, it can also be applied to pose estimation using RGB-D inputs.
This is because it estimates the pose based on correspondences between two images.
To achieve this, in the coarse estimation stage, we use semi-dense correspondences and employ the Kabsch algorithm \cite{kabsch1976solution} instead of PnP \cite{Lepetit2009ep} for pose estimation.
In the refinement stage, instead of returning the estimated pose from the differentiable PnP layer, we estimate the pose using the model’s outputs: dense correspondences and certainty.
Certainty represents the degree to which the flow from the rendered image reaches the object surface of the query image; therefore, we use only correspondences with a certainty of 0.5 or higher, thereby excluding the influence of occluded correspondences. 
To obtain the final pose, we utilize MAGSAC++ \cite{barath2019magsacplusplus} in our implementation.

Table \ref{tab:supp_results} presents our pose estimation results on the seven core datasets of the BOP challenge using RGB-D inputs.
GenFlow \cite{moon2024genflow}, the best overall method in the 2023 BOP challenge, reported results only for CNOS \cite{nguyen2023cnos} detection, while FoundationPose \cite{wen2024foundationpose} reported results solely for SAM-6D \cite{Lin2024sam6d} detection.
To ensure a fair comparison, we conducted separate experiments for each detection method and reported our results individually.
Additionally, we included the submission results that achieved the highest scores among various experimental settings for each compared method.
Specifically, we cite MegaPose \cite{labbe2022megapose} results using 10 hypotheses and Teaser++ \cite{Yang20tro-teaser}, and GenFlow results using 16 hypotheses.

\subsection{Pose Selection Ablation}
\def\angle{70}

\begin{table}[t!]
    \setlength{\tabcolsep}{2.5pt}
    \scriptsize %
    \begin{center}
    \begin{tabularx}{1.0\linewidth}{r l |Y Y Y Y Y Y Y |Y}
    \toprule
    \# \vspace{0.4ex} &
    Method &
    \rotatebox[origin=lB]{\angle}{LM-O} &
    \rotatebox[origin=lB]{\angle}{T-LESS} &
    \rotatebox[origin=lB]{\angle}{TUD-L} &
    \rotatebox[origin=lB]{\angle}{IC-BIN} &
    \rotatebox[origin=lB]{\angle}{ITODD} &
    \rotatebox[origin=lB]{\angle}{HB} &
    \rotatebox[origin=lB]{\angle}{YCB-V} &
    \rotatebox[origin=lB]{\angle}{Mean} \\
    \toprule
    
    \newtag{1}{supp_ab.1} & Ours (Proposed) & 65.5	& 64.8	& 72.9	& 54.2	& 49.0 & 84.9 & 68.9	& 65.7	\\

    \newtag{2}{supp_ab.2} & Ours (Same as GenFlow) & 65.0	& 64.7	& 73.0	& 53.5	& 48.7 & 84.8 & 69.0	& 65.5	\\
    
    \newtag{3}{supp_ab.3} & GenFlow \cite{moon2024genflow} & 65.1	& 64.5	& 72.6	& 53.0	& 47.5 & 84.7 & 69.0 & 65.2	\\

    \bottomrule
    \end{tabularx}
    \end{center}
    \caption{\textbf{Selection Model Ablation Study.} Ablation study on the impact of positive rotation thresholds and backbone in pose selection models. 
    }
    \label{tab:sel_ablations}
\end{table}

To enhance the pose selection model’s ability to identify poses close to the ground truth, we used a smaller rotation threshold when defining positives and negatives.
We conducted comparative experiments to verify this approach, as shown in Table \ref{tab:sel_ablations}.

GenFlow reuses the coarse estimation model for pose selection, employing a rotation threshold of 15 degrees.
For comparison, we trained a pose selection model with a ConvNeXt \cite{liu2022convnet} architecture, following GenFlow’s settings (row \ref{supp_ab.3}).
We also trained our model using the same configuration as GenFlow (row \ref{supp_ab.2}).
Comparing these models with our proposed pose selection model (row \ref{supp_ab.1}), we confirm that a tighter positive threshold improves accuracy in precise pose estimation.

\section{Qualitative Results}

\subsection{Coarse Estimation}
\begin{figure*}[t]
	\centering
	\includegraphics[width=0.95\linewidth]{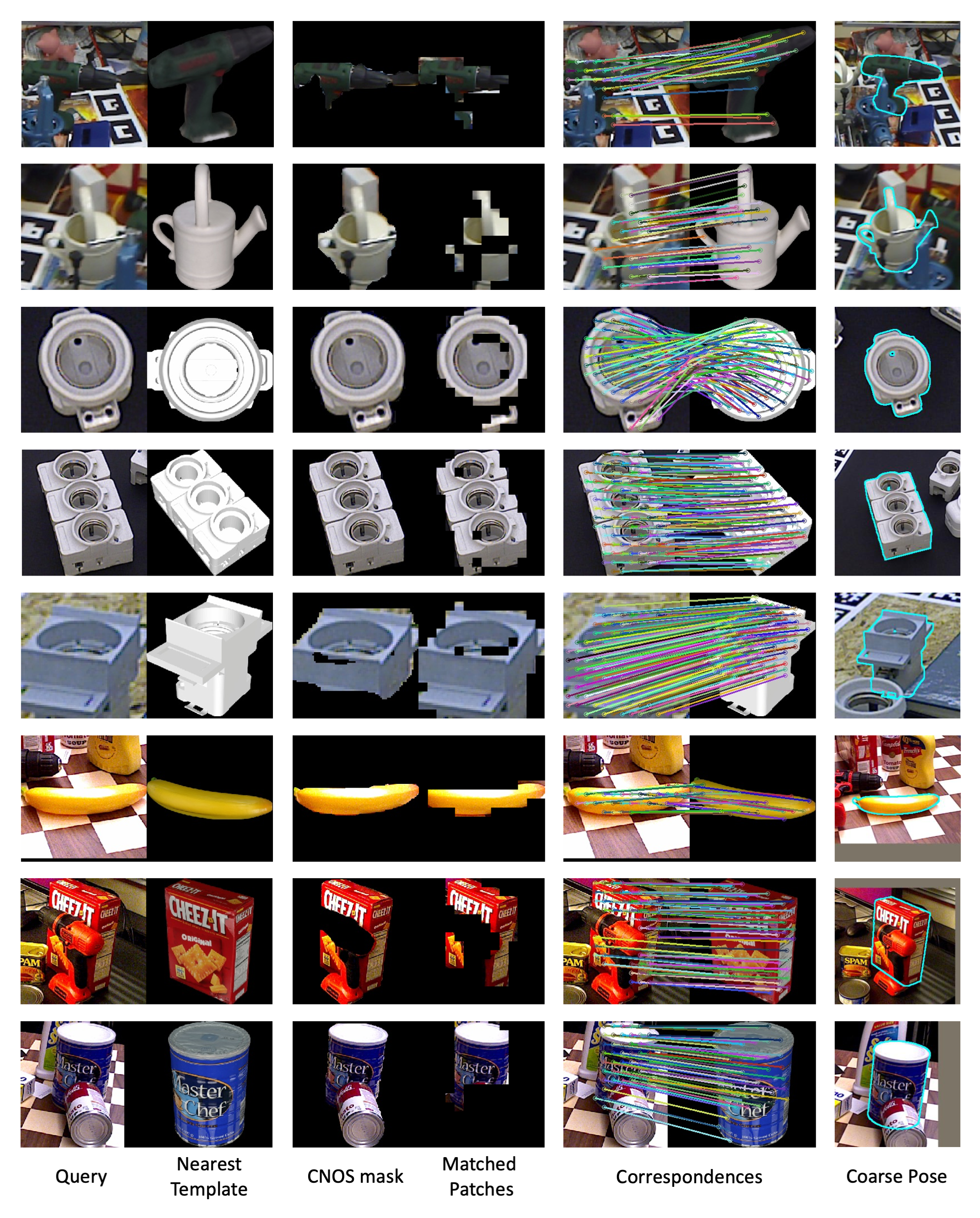}

	\caption{{\bf Qualitative Results of Coarse Estimation.} The first two columns on the left display the model’s query image and the template with the highest similarity score to the query image. The third and fourth columns compare the CNOS \cite{nguyen2023cnos} segmentation mask with patches that the model did not classify as 'no-match'. From the fifth to the last columns, the correspondences between the query image and the template, as well as the resulting pose estimation results, are shown.
    }
	\label{fig:supp_coarse}
\end{figure*}

Fig. \ref{fig:supp_coarse} presents the qualitative results of the coarse estimation. 
From left to right, the first and second columns display the query image and the template with the highest similarity score, respectively.
The third and fourth columns display the segmentation mask from CNOS \cite{nguyen2023cnos} and the patches that our model did not classify as "no match."
From the fifth to the final columns, we present the semi-dense correspondences between the query image and the most similar template, along with the estimated poses derived from them.
We can observe that our proposed method accurately estimates correspondences across various datasets, including YCB-V \cite{xiang2018posecnn}, which consists of texture-rich objects, and T-LESS \cite{hodan2017tless}, which consists of low-texture objects.

As demonstrated in the third and fourth rows, our proposed method is robust to detection errors. 
Segmentation masks from CNOS or SAM-6D \cite{Lin2024sam6d} rely on SAM \cite{kirillov2023segment}, which does not consider object information, leading to over-segmentation (see row 5) or under-segmentation (see row 8).
As reported in the ablation experiments of FoundPose \cite{ornek2024foundpose}, such segmentation errors can significantly affect pose estimation accuracy.
In contrast, our model—similar to detector-free methods in correspondence estimation—jointly considers the query image and the template.
This approach allows our model to estimate poses robustly against detection errors.

\subsection{Pose Refinement}
\begin{figure*}[t]
	\centering
	\includegraphics[width=0.95\linewidth]{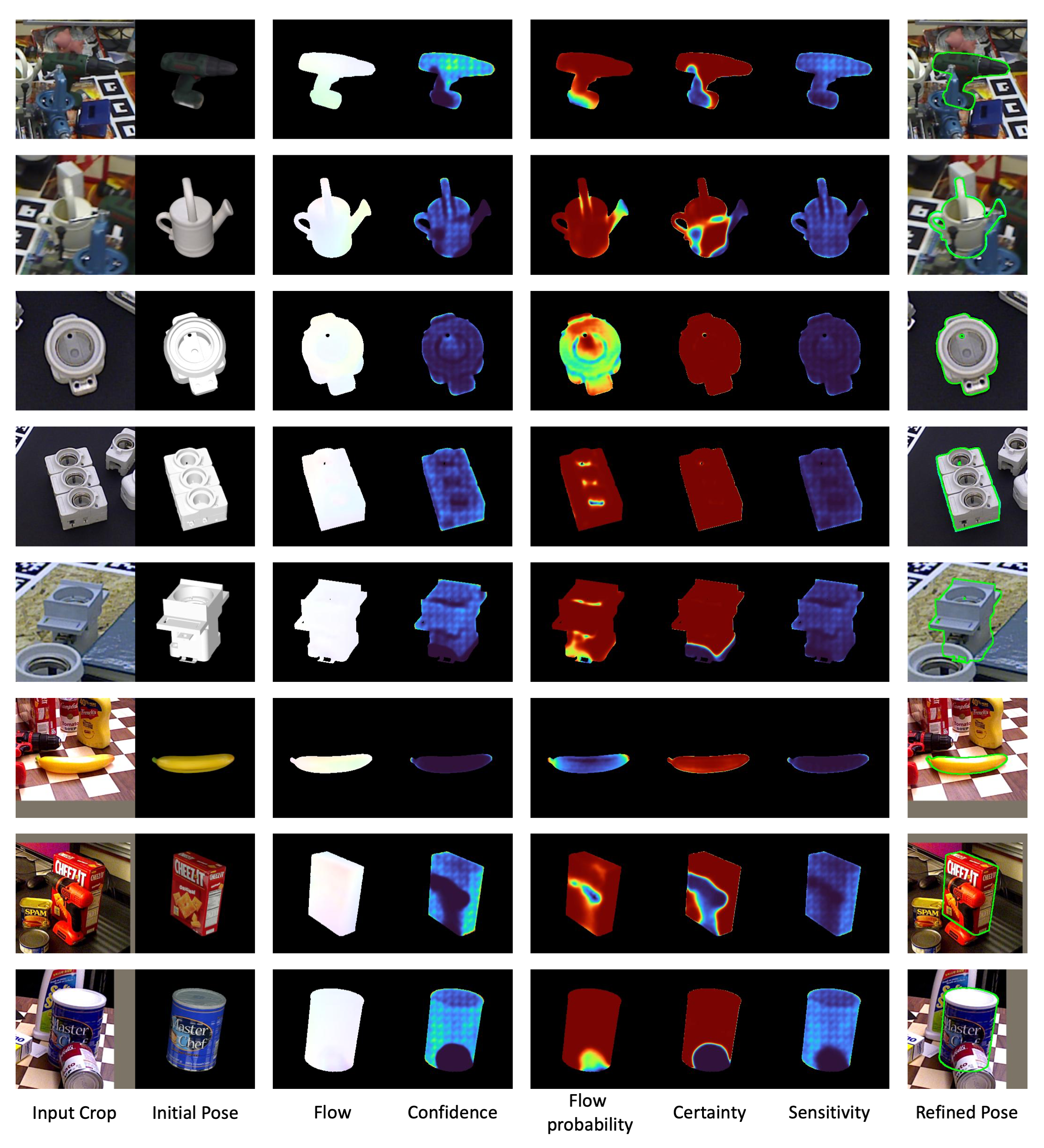}

	\caption{{\bf Qualitative Results of Pose Refinement.} From left to right: query image, initial pose rendering, flow, confidence, flow probability, certainty, sensitivity, and the refined pose (legend: 0.0~ \includegraphics[height=1.8mm]{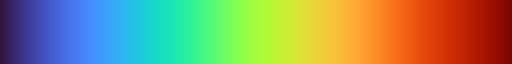} 1.0~). The flow probability and certainty reduce confidence in ambiguous or occluded areas, while sensitivity increases confidence in textured regions and object edges to improve pose refinement. 
    }
	\label{fig:supp_refine}
\end{figure*}

Fig. \ref{fig:supp_refine} presents a visualization of the pose refinement process. 
From left to right, the first two columns display the query image and the rendered image of the initial pose, respectively.
The third and fourth columns show the visualization of flow and confidence, which are inputs to the differentiable PnP layer.
The fifth to seventh columns show the flow probability, certainty, and sensitivity, which are used to calculate the confidence.
The last column shows the model’s output—the refined pose of the object.

As seen in the fifth and sixth rows, the flow probability lowers the confidence in ambiguous areas such as the object’s self-occlusion.
Similarly, the certainty reduces confidence in areas where it is difficult to trust correspondences because of occlusions. Conversely, the sensitivity increases confidence in regions with rich texture (see the eighth row) or at the object’s edges (see the sixth row) to achieve accurate pose refinement.

\label{sec:add_exp}


\end{document}